\documentclass[12pt]{article}

\usepackage{amssymb}
\usepackage{amsmath}
\usepackage{amsfonts}
\usepackage{amsthm}
\usepackage{bbm}
\usepackage[margin=1in]{geometry}
\usepackage{graphicx}
\usepackage{natbib}
\usepackage{color}
\usepackage{url}
\usepackage{booktabs}

\usepackage[font=small,labelfont=bf]{caption}
\usepackage{color}
\usepackage{mathtools}
\usepackage{bbm}
\usepackage{subcaption}
\usepackage{multirow}
\usepackage{xcolor}

\newcommand{\CV}{\mathrm{CV}}
\newcommand{\E}{\mathrm{E}}

\newcommand{\Cov}{\mathrm{Cov}}

\newcommand{\argmin}{\mathrm{argmin}}

\newtheorem{myexample}{Example}

\theoremstyle{plain}
\newtheorem{lemma}{Lemma}
\newtheorem{theorem}{Theorem}

\usepackage[affil-it]{authblk} 
\usepackage{etoolbox}
\usepackage{lmodern}

\makeatletter
\patchcmd{\@maketitle}{\LARGE \@title}{\fontsize{16}{19.2}\selectfont\@title}{}{}
\makeatother

\title{Selective prediction-set models with coverage rate guarantees}
\author{Jean Feng$^{1}$, Arjun Sondhi$^{2}$, Jessica Perry$^3$, and Noah Simon$^3$}
\affil{Department of Epidemiology and Biostatistics, University of California, San Francisco$^{1}$,\\ Flatiron Health$^{2}$,\\	Department of Biostatistics, University of Washington$^{3}$\\}

\date{}

\begin{document}
\maketitle
	\begin{abstract}
		{
The current approach to using machine learning (ML) algorithms in healthcare is to either require clinician oversight for every use case or use their predictions without any human oversight.
We explore a middle ground that lets ML algorithms abstain from making a prediction to simultaneously improve their reliability and reduce the burden placed on human experts.
To this end, we present a general penalized loss minimization framework for training selective prediction-set (SPS) models, which choose to either output a prediction set or abstain.
The resulting models abstain when the outcome is difficult to predict accurately, such as on subjects who are too different from the training data, and achieve higher accuracy on those they \textit{do} give predictions for.
We then introduce a model-agnostic, statistical inference procedure for the coverage rate of an SPS model that ensembles individual models trained using K-fold cross-validation.
We find that SPS ensembles attain prediction-set coverage rates closer to the nominal level and have narrower confidence intervals for its marginal coverage rate.
We apply our method to train neural networks that abstain more for out-of-sample images on the MNIST digit prediction task and achieve higher predictive accuracy for ICU patients compared to existing approaches.
		}
	\end{abstract}

\section{Introduction}

With the rapid development of large medical datasets, a growing number of clinical prediction models have been developed using sophisticated machine learning (ML) algorithms like neural networks (NNs) and tree-based methods.
Unlike simpler parametric approaches,  flexible ML algorithms can learn nonlinearities and interactions in the data to achieve unprecedented performance in complex problem domains.
Indeed, a growing number of ML-based software-as-a-medical devices (SaMDs) have now gained regulatory approval to analyze medical images, electronic health records, and more \citep{Benjamens2020-eh}.

Nevertheless, the safety of these algorithms continues to be a concern for a variety of reasons, including a lack of theoretical guarantees and issues of model interpretability \citep{Challen2019-zn}.
This is evident in the regulatory approval patterns by the U.S FDA  for ML-based SaMDs.
Currently, the handful of ML-based SaMDs approved for use without \textit{any} human supervision are limited to low-risk settings \citep{Office_of_the_Commissioner2018-aa, Muehlematter2021-iq}.
In all other cases, clinicians are expected to oversee their \textit{every} use to ensure their safety and effectiveness.
This work explores the middle ground, in which the ML algorithm provides guidance for ``easy'' cases and appeals to the human expert in the remaining ``difficult'' cases.
Models that have this ability to abstain from giving a prediction are also known as selective prediction models \citep{Chow1970-uc,Tortorella2000-yc,Bartlett2008-sn, El-Yaniv2010-kd}.
For example, an ML algorithm may risk stratify patients with simple medical histories and unambiguous risk profiles; the clinician would analyze the remaining, more complex cases with multiple comorbidities.
Another example is predicting the length-of-stay (LOS) for patients, which can help hospital managers schedule elective surgeries, identify high-risk patients, manage hospital capacity, and forecast hospital budget \citep{Robinson1966-wu, ettema2010prediction, zimmerman2006intensive, verburg2017models}.
It can be nearly impossible to get meaningful LOS predictions for certain patients due to a lack of information or complexity of their medical history; in such cases, the model output would be useless or misleading.
Instead, the hospital manager may collect more information from the patient or clinician to get a usable prediction or wait until the patient is admitted to get a refined estimate \citep{Robinson1966-wu}.
Through this collaboration between practitioners and the ML algorithm, we hope to simultaneously reduce the workload for the former and address safety concerns for the latter.

To this end, the first part of this paper presents a new penalized loss minimization framework for training selective prediction-set (SPS) models, which choose to either output a prediction set to quantify uncertainty or abstain from outputting any prediction.
Our general framework is compatible with any differentiable model class; many of our examples fit NNs.
In contrast to prior selective prediction methods that either output point estimates or abstain, we use prediction \textit{sets} to better quantify uncertainty and inform patient expectations \citep{Kosel2019-ji}.
Recent works aim to improve uncertainty quantification in NNs using conditional density estimation \citep{Lakshminarayanan2017-wn} and quantile estimation \citep{Taylor2000-iu}.
Nevertheless, these methods can still be overconfident for ``outliers.''
An ad-hoc solution is to train an outlier detector separately \citep{Pimentel2014-uw}.
Here we consider training the prediction and decision function \textit{simultaneously} so that the prediction model will optimize performance on the ``easy'' cases.

Although our penalized estimator attains the nominal conditional coverage rate asymptotically, the trained model may not necessarily achieve this due to finite sample sizes and the nonconvex training objective.
As such, the second part of this paper presents a statistical inference procedure for the average coverage rate of an SPS model trained using our proposed penalization framework or, more generally, any black-box estimation technique.
Existing model-agnostic statistical inference procedures rely on a single training-validation split, reserving a portion of data for model evaluation \citep{Platt1999-yw,Zadrozny2002-ro,Kuleshov2015-yg,Guo2017-gy}.
To use the data more fully, we introduce a $K$-fold cross-validation (CV) procedure that constructs an ensemble of SPS models using the $K$ individual models and performs statistical inference for the marginal prediction-set coverage rate.
Compared to the individual models, the ensemble attains coverage rates closer to the nominal level and the confidence interval for its marginal coverage rate is narrower.
We note that the conformal inference methods also aim to construct prediction sets with coverage guarantees \citep{Vovk2005-wf, Sadinle2019-ph}, but focus on finite-sample guarantees and do not include an abstention option.

\begin{figure}
\centering
\includegraphics[width=0.6\linewidth]{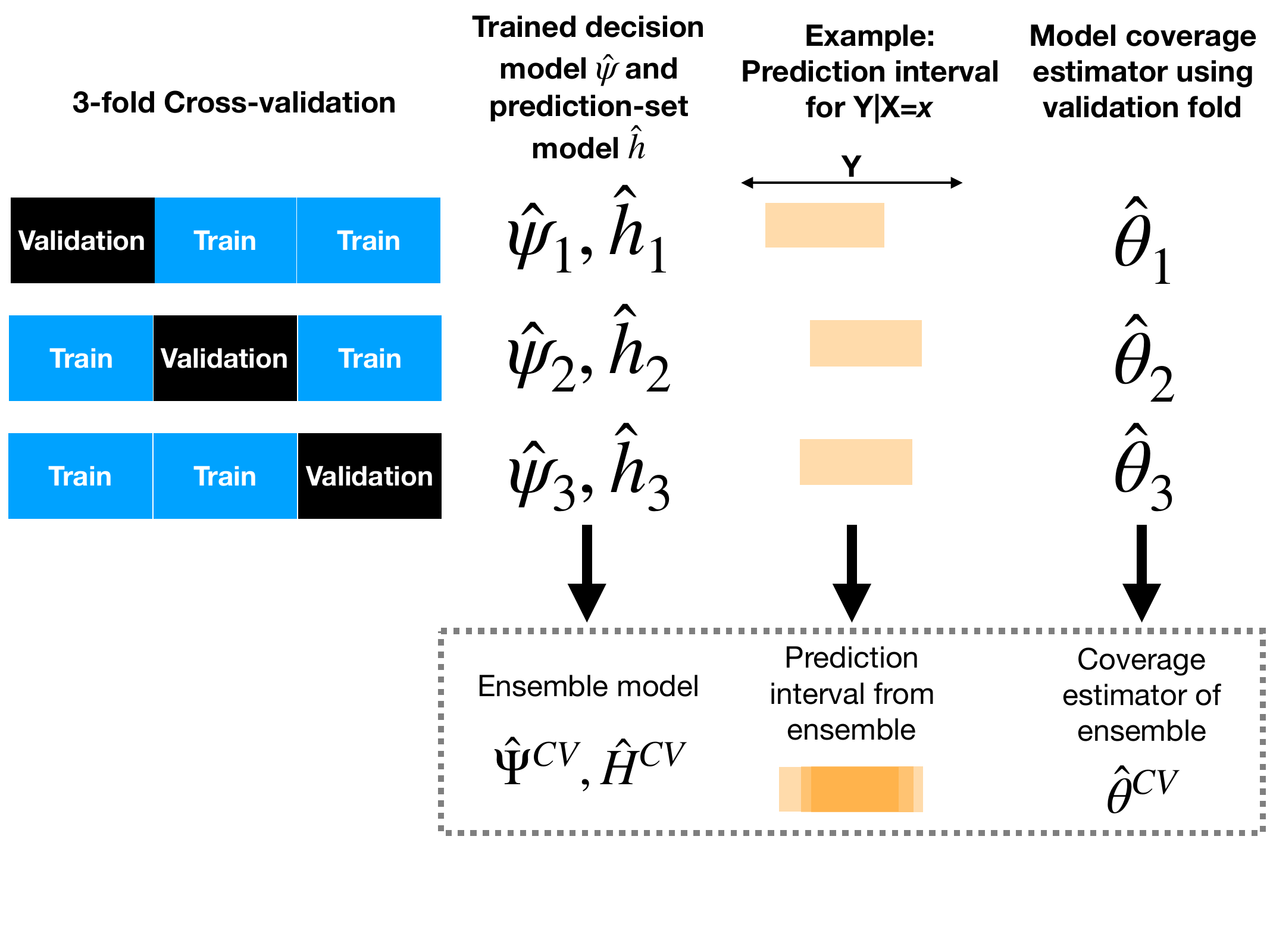}
\caption{
Procedure for training and calibrating an ensemble of SPS models using $K$-fold CV with $K = 3$.
For each $k$, perform penalized estimation for SPS models on all but the $k$th fold to train decision model $\hat{\psi}_k$ and prediction-set model $\hat{h}_k$.
As an example, we illustrate estimated prediction intervals for $Y | X = x$ for a given $x$.
The estimators of the marginal prediction-set coverage rates of the $k$th model, denoted $\hat{\theta}_k$, are computed using the $k$th fold.
We then combine the results from all $K$ folds as follows.
The ensemble model is a uniform mixture of the trained models.
We visualize its output by overlaying estimated prediction intervals from the individual models.
We can estimate the coverage rate of the ensemble model by combining coverage estimates of the individual models.
This figure appears in color in the electronic version of this article.
}
\label{fig:summary_everything}
\end{figure}

A visual summary of our final procedure, from training the SPS model to inference of its coverage rate, is given in Figure~\ref{fig:summary_everything}.
The rest of this paper is organized as follows.
Section~\ref{sec:setup} provides the necessary background and notation for SPS models.
Section~\ref{sec:dec-pred} presents \textbf{p}enalized estimation for \textbf{S}elective \textbf{P}rediction-\textbf{S}et models (pSPS), in which we present two new penalty functions.
In Section~\ref{sec:calibration}, we describe a new statistical inference procedure, \textbf{c}ross-validation for \textbf{S}elective \textbf{P}rediction-\textbf{S}et models (cSPS), to calibrate prediction-set coverage rates.
We present simulation results and empirical analyses on the MNIST digit classification task in Sections~\ref{sec:simulations} and \ref{sec:mnist}, respectively.
Finally, Section \ref{sec:mimic} is a real-world case study on predicting intensive care unit (ICU) mortality and visit length in MIMIC-III critical care database \citep{Johnson2016-gu}.

\section{Background and notation: SPS models}
\label{sec:setup}

Consider a prediction problem with covariates $X$ from a closed sample space $\mathcal{X} \subset \mathbb{R}^p$ and a categorical or continuous outcome $Y$ with sample space $\mathcal{Y} \subseteq \mathbb{R}$.
Let $p^*(x)$ be the true density of $X$, $P$ the joint distribution of $(X,Y)$, and $S(\mathcal{Y})$ the collection of all subsets of $\mathcal{Y}$.


An SPS model is defined by a decision function $\psi: \mathcal{X} \mapsto \{0, 1\}$ and prediction function $h: \mathcal{X} \mapsto S(\mathcal{Y})$.
When $\psi(x) = 0$, the model abstains for $x$; when $\psi(x) = 1$, the model accepts and outputs the prediction set $h(x)$.
For continuous outcomes, we fit $h$ to output contiguous prediction intervals (PIs).
For any $\alpha \ge 0$ and $x \in \mathcal{X}$, a set $\tilde{h}$ is a $1 - \alpha$-prediction set for $Y|X = x$ if its conditional coverage $\Pr\left(Y \in \tilde{h} | X = x\right)$ is at least $1 - \alpha$.

Given $n$ independent and identically distributed (iid) observations from $P$, we would like to design an estimation procedure $(h_n, \psi_n)$ that achieves \textit{uniform} $1-\alpha$-conditional coverage.
That is, we would like the trained model $(\hat{h}_n, \hat{\psi}_n)$ to satisfy

\begin{align}
\Pr\left(Y \in \hat{h}_n(x) \mid X = x \right) \ge 1 - \alpha 
\label{eq:unif_cov}
\end{align}
for all accepted $x \in \mathcal{X}$ (i.e. $ \hat{\psi}_n(x)=1$), where the probability in \eqref{eq:unif_cov} is with respect to the $n$ observations and a test observation $Y|X = x$.
However, \citet{Lei2014-uh} and \citet{Barber2019-fr} showed that the only procedure that satisfies \eqref{eq:unif_cov} in finite samples for all distributions $P$ for continuous $Y$ is the trivial solution that returns infinite-length prediction intervals almost everywhere.
Given that this impossibility result, we consider a two-part solution: first, an estimation procedure to fit an SPS model $(h_n, \phi_n)$ that attains uniform conditional coverage rates asymptotically and second, a statistical inference procedure for its marginal coverage rate with respect to the accepted observations.

\section{Penalized estimation for SPS models}
\label{sec:dec-pred}

Here, we present a penalized empirical loss minimization framework for learning SPS models. We refer to this procedure as penalized estimation for selective prediction-set models (pSPS), and it can be applied using either a decision-theoretic or robust maximum-likelihood estimation (MLE) approach. Let the training observations $(X_i, Y_i)$ for $i = 1, 2, ..., n$ be drawn iid from a joint distribution $P$.
The decision-theoretic approach learns the decision function $\psi$ and prediction-set function $h$ by solving
\begin{align}
	\min_{h, \psi}
	\underbrace{
		\frac{1}{n} \sum_{i=1}^n
		\left\{
		\ell_{\alpha}(h(X_i), Y_i)
		\psi(X_i)
		+ \delta (1 - \psi(X_i))
		\right\}
	}_\text{Adaptively truncated loss} +
	\underbrace{\lambda_0
		\frac{1}{n} \sum_{i=1}^n
		\ell_{\alpha}(h(X_i), Y_i)
	}_\text{Information borrowing penalty} +
	\underbrace{
		\lambda_1 \int_{\mathcal{X}}  \psi(x) dx.
	}_\text{Uniform acceptance penalty}
	\label{eq:obj_dt}
\end{align}
The robust MLE approach learns the decision function $\psi$ and conditional density function $f$ by solving
\begin{align}
	\min_{f, \psi}
	\underbrace{
		\frac{1}{n} \sum_{i=1}^n
		\left\{
		- \log f(Y_i | X_i)
		\psi(X_i)
		+ \delta (1 - \psi(X_i))
		\right\}
	}_\text{Adaptively truncated loss} +
	\underbrace{\lambda_0
		\frac{1}{n} \sum_{i=1}^n
		- \log f(Y_i | X_i)
	}_\text{Information borrowing penalty} +
	\underbrace{
		\lambda_1 \int_{\mathcal{X}}  \psi(x) dx.
	}_\text{Uniform acceptance penalty}
	\label{eq:obj_mle}
\end{align}
The prediction-set function $h$ is then constructed based on the estimated conditional density function. In both cases, the objective function is the sum of three components: an adaptively truncated empirical loss, an information borrowing penalty scaled by penalty parameter $\lambda_0 \ge 0$, and a uniform acceptance penalty scaled by penalty parameter $\lambda_1 \ge 0$.

The \textbf{adaptively truncated empirical loss} incurs the prediction loss if the model accepts and the cost of abstention $\delta$ if the model abstains.
In the decision-theoretic approach, the prediction loss maps prediction sets to a real-valued loss via the function $\ell_{\alpha}: \mathcal{S}(\mathcal{Y}) \times \mathcal{Y} \mapsto \mathbb{R}$; as indicated by the subscript $\alpha$, the definition of the prediction-set loss varies based on the desired coverage rate.
In the robust MLE approach, the prediction loss is the negative log likelihood of $Y|X$, which does not vary with $\alpha$.
This prediction loss must be balanced with the cost of abstention $\delta$.
Ideally, $\delta$ is chosen based on prior knowledge.
For example, if the plan is to query a human expert whenever the ML algorithm abstains, $\delta$ should correspond to the cost of querying a human expert relative to the prediction loss.
Nevertheless, it can be difficult to choose $\delta$ upfront and we discuss alternative selection strategies in Section~\ref{sec:model_impl}.

The \textbf{information borrowing penalty} is the usual empirical prediction loss over \textit{all} observations.
As such, it encourages the model to learn from observations that it has abstained on.
To make this more concrete, suppose we trained an SPS model using the above objective with $\lambda_0 = 0$.
If this fitted model abstains for 50\% of the training observations, the prediction model has effectively trained on only half of the available data.
However, the remaining observations likely provide useful information for improving prediction accuracy within the acceptance region.
By tuning the penalty parameter $\lambda_0$, we can modulate how much information is borrowed from the abstention region.

Finally, the \textbf{uniform acceptance penalty} penalizes all acceptances equally by integrating the decision function with respect to the Lebesgue measure over the entire domain  $\mathcal{X}$.
Thus it encourages the model to abstain for unfamiliar observations.
Our choice to use the Lebesgue measure rather than the measure of $X$ is deliberate.
The latter would assign a higher penalty to accepting inputs with \textit{higher} density, and in the extreme case where $p^*(x) = 0$ would assign no penalty, the opposite of the desired behavior.

When these three components are combined in pSPS, the trained model decides whether or not to abstain for a given input query $x$ based on the expected loss at $x$, the density at $x$, and the penalty parameter $\lambda_1$.
It is encouraged to abstain if the loss is high, $x$ is an outlier and belongs to a minority population (i.e. $p^*(x)$ is small but nonzero), or $x$ is out-of-distribution (i.e. $p^*(x) = 0$).
We formalize this behavior in Theorem~\ref{thrm:population_version} below by analyzing the population-level SPS model with respect to a fully nonparametric model class.
We show that they attain uniform $1-\alpha$ conditional coverage, asymptotically.
Because the population-level SPS model is the same in this setting whether we optimize over binary decision functions or continuous functions mapping to values between 0 and 1, we employ this relaxation in practice.

\begin{theorem}
	Let $\mathcal{D}$ be the set of all functions $\psi: \mathcal{X} \mapsto [0,1]$.

	\textit{Decision-theoretic approach:}
	Let $\mathcal{H}$ be the set of all functions $h: \mathcal{X} \mapsto \mathcal{S} \subseteq S(\mathcal{Y})$.
	Define the function $\ell^*_{\alpha}:\mathcal{X} \mapsto \mathbb{R}$ as $
	\ell^*_{\alpha}(x) = \min_{\tilde{h} \in \mathcal{S}}
	E \left[\ell_{\alpha}\left(\tilde{h}, Y\right) \mid X = x \right ]$ and suppose $\ell^*_{\alpha}(x)$ exists for all $x$.
	For any $\delta \in \mathbb{R}$ and $\lambda_0, \lambda_1 > 0$, the population-level SPS model
	\begin{align}
		\psi_{\alpha, \delta, \lambda_0, \lambda_1}, h_{\alpha, \delta, \lambda_0, \lambda_1}
		\in \argmin_{\psi \in \mathcal{D}, h \in \mathcal{H} }
		\E\left[
		\ell_{\alpha}\left( h(X), Y \right) (\psi(X) + \lambda_0) + \delta (1 - \psi(X))
		\right]
		+\lambda_1 \int_{\mathcal{X}} \psi(x) dx
		\label{eq:dec_pop}
	\end{align}
	satisfies for almost every $x \in \mathcal{X}$
	\begin{align}
		\psi_{\alpha, \delta, \lambda_0, \lambda_1}(x)
		&=
		\begin{cases}
		1 & \text{if }  p^*(x) \left(\ell^*_{\alpha}(x) - \delta\right)
		< -\lambda_1\\
		0 & \text{if }  p^*(x) \left(\ell^*_{\alpha}(x) - \delta\right)
		> -\lambda_1
		\end{cases}
		\label{eq:pop_dec}
		\\
		h_{\alpha, \delta, \lambda_0, \lambda_1}(x)
		& \in
		\argmin_{\tilde{h} \in \mathcal{S}}
		E\left[\ell_{\alpha}\left(\tilde{h}, Y\right) \mid X = x \right ]
		\forall  x \text{ s.t. } p^*(x) > 0.
		\label{eq:pop_set}
	\end{align}
	\textit{Robust MLE approach:} Let $\mathcal{F}$ be the set of all possible density functions and $\mathcal{Q}$ be the set of functions mapping $\mathcal{X}$ to $\mathcal{F}$.
	For any $\delta \in \mathbb{R}$ and $\lambda_0, \lambda_1 > 0$, the population-level selective conditional density model
	\begin{align}
		\psi_{\delta, \lambda_0, \lambda_1}, f_{\delta, \lambda_0, \lambda_1}
		\in \argmin_{\psi \in \mathcal{D}, f \in \mathcal{Q} }
		\E\left[
		- \log f(Y|X) (\psi(X) + \lambda_0) + \delta (1 - \psi(X))
		\right]
		+\lambda_1 \int_{\mathcal{X}} \psi(x) dx
		\label{eq:model_pop_loglik}
	\end{align}
	satisfies for almost every $x \in \mathcal{X}$ and $ y\in \mathcal{Y} $
	\begin{align}
		{\psi}_{\delta, \lambda_0, \lambda_1}(x)
		&=
		\begin{cases}
		1 & \text{if } p^*(x)
		\left(
		-E_{Y|X=x}[\log p^*(Y|x)] - \delta
		\right)
		< -\lambda_1\\
		0 & \text{if } p^*(x)
		\left(
		-E_{Y|X=x}[\log p^*(Y|x)] - \delta
		\right)
		> -\lambda_1
		\end{cases}
		\label{eq:pop_rmle}
		\\
		{f}_{\delta, \lambda_0, \lambda_1}(y|x)
		& = p^*(y|x)
		\qquad \forall x, y \text{ s.t. } p^*(x) > 0.
		\label{eq:rmle_pop_set}
	\end{align}
	\label{thrm:population_version}

\end{theorem}
\noindent
The above result shows that the population-level decision models for both approaches have similar forms and abstain for all $x$ with $p^*(x) = 0$ as long as $\lambda_1 > 0$.
For $x$ with small $p^*(x)$, the former abstains based on the expected prediction set loss and the latter depends on the differential entropy for $Y|X = x$.

Theorem~\ref{thrm:population_version} highlights the difference between simultaneously training the decision and prediction-set functions versus separately training the decision function using an outlier detection method.
The population-level pSPS model accepts an observation $x$ only if its expected loss is acceptably small, where the maximum acceptable expected loss at $x$ is proportional to the density $p^*(x)$.
On the other hand, an outlier detector abstains for all $x$ where $p^*(x)$ is below some threshold.
Therefore, pSPS outperforms a separately-trained outlier detector for $x$ in regions with low-density but where outcomes can be predicted with high confidence.

Interestingly, $\lambda_0$ has no consequential effect on the population-level SPS model, as seen in Theorem~\ref{thrm:population_version}.
Nevertheless, we find that the information borrowing penalty can drastically improve performance in finite samples.

\subsection{Comparing the decision-theoretic and robust MLE approaches}
\label{sec:examples}


If the outcome is continuous, we estimate PIs using 1) the decision-theoretic approach and optimize over a nonparametric class that specifies the interval boundaries or 2) the robust MLE approach and optimize over a parametric family of conditional distributions. The choice between approaches reduces to the typical bias-variance tradeoff when choosing between nonparametric versus parametric models. That is, the former is suitable when there is enough data for fitting an accurate nonparametric model, whereas the latter is advantageous with smaller datasets and correctly specified models (or model misspecification is not severe).
In fact, we can show that two approaches yield the same population-level SPS models under certain conditions (see Theorem~\ref{thrm:equiv_models} and Example~\ref{example:gauss} of the Appendix).
For concreteness, we present an example for each approach below.

\begin{myexample}[Decision-theoretic approach]
\label{ex:continuous_dt}
For interval $h$, with center $\mu_h$ and radius $r_h$, and outcome $y\in\mathcal{Y}$, let the prediction-set loss be the \textit{absolute discrepancy} \citep{Rice2008-ia}
\begin{align}
\ell_{\alpha}(h, y) = \alpha r_{h} +
\underbrace{
	\left ((\mu_h - r_{h}) - y \right )^+
	+ \left (y - (\mu_{h} + r_{h}) \right )^+
}_{
	\text{Distance outside interval}
}.
\label{eq:abs_loss}
\end{align}
It is straightforward to show the population-level prediction-interval function \eqref{eq:pop_set} is the $\alpha/2$ and $1 - \alpha/2$ quantiles of $Y|X=x$, which achieves uniform $1 - \alpha$ conditional coverage.
\end{myexample}

\begin{myexample}[Robust MLE approach]
	\label{ex:continuous_mle}
	We constrain the conditional density function such that the density function $f(\cdot)$ is from a parametric family, such as the Gaussian distribution.
	If the true model is not within the parametric family, the model is misspecified.
	To protect against this, one solution is to optimize over a sufficiently rich class of decision functions, since it can learn to abstain in the misspecified regions.
\end{myexample}

For categorical outcomes, we recommend using the robust MLE approach, in which we parameterize $Y|X$ as a multinomial distribution.
Moreover, we can avoid issues of model misspecification by estimating the conditional probabilities using a semi-/non-parametric model.
The difficulty in taking the decision-theoretic approach is that there is no suitable prediction-set losses for categorical outcomes to our knowledge.
Although the step function loss \citep{Rice2008-ia} may seem like a good candidate, it cannot be used directly because it requires knowing $p(Y|x)$.

\subsection{Model implementation}
\label{sec:model_impl}

\begin{figure}
\centering
\includegraphics[width=0.45\linewidth]{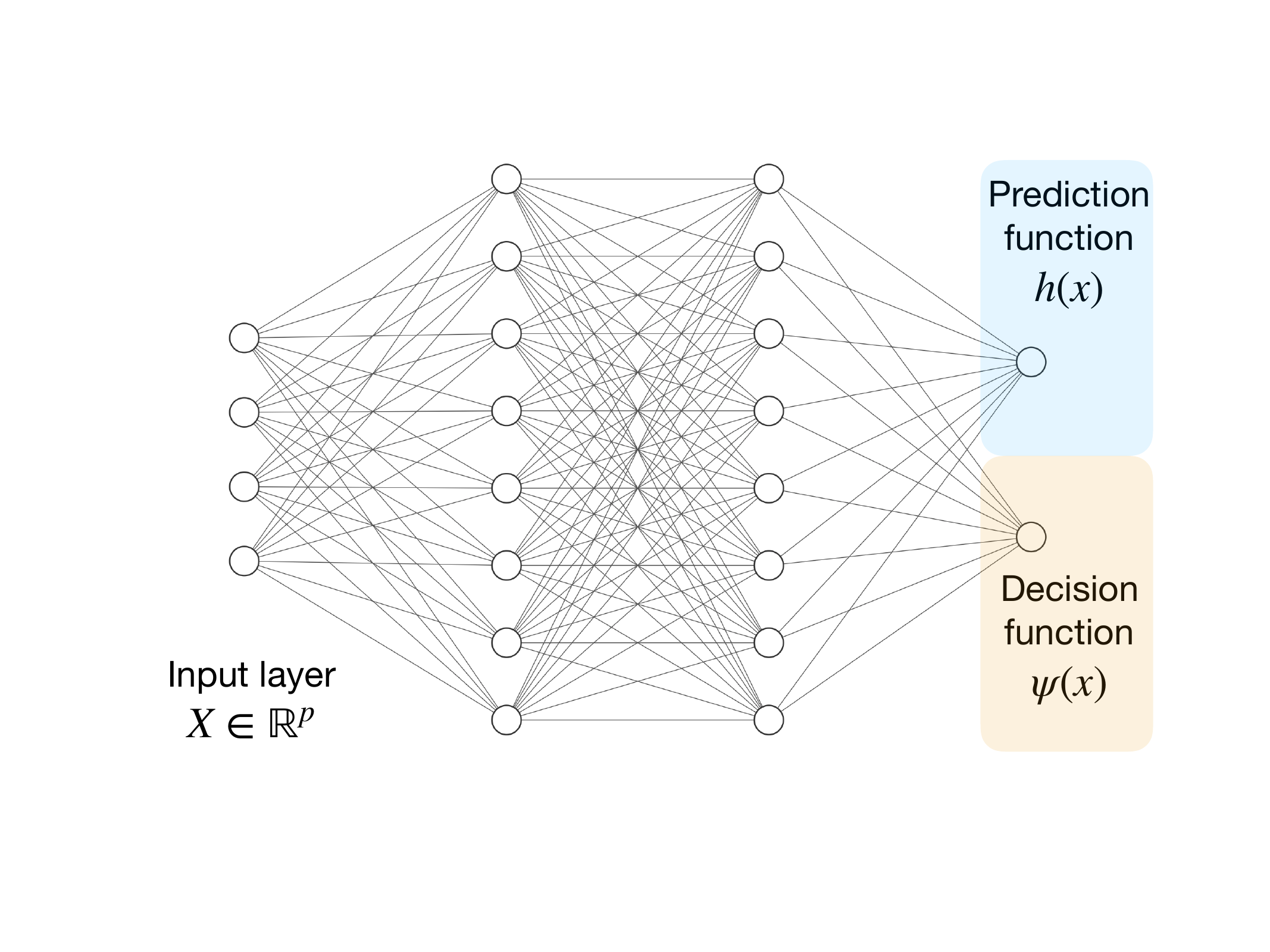}
\includegraphics[width=0.45\linewidth]{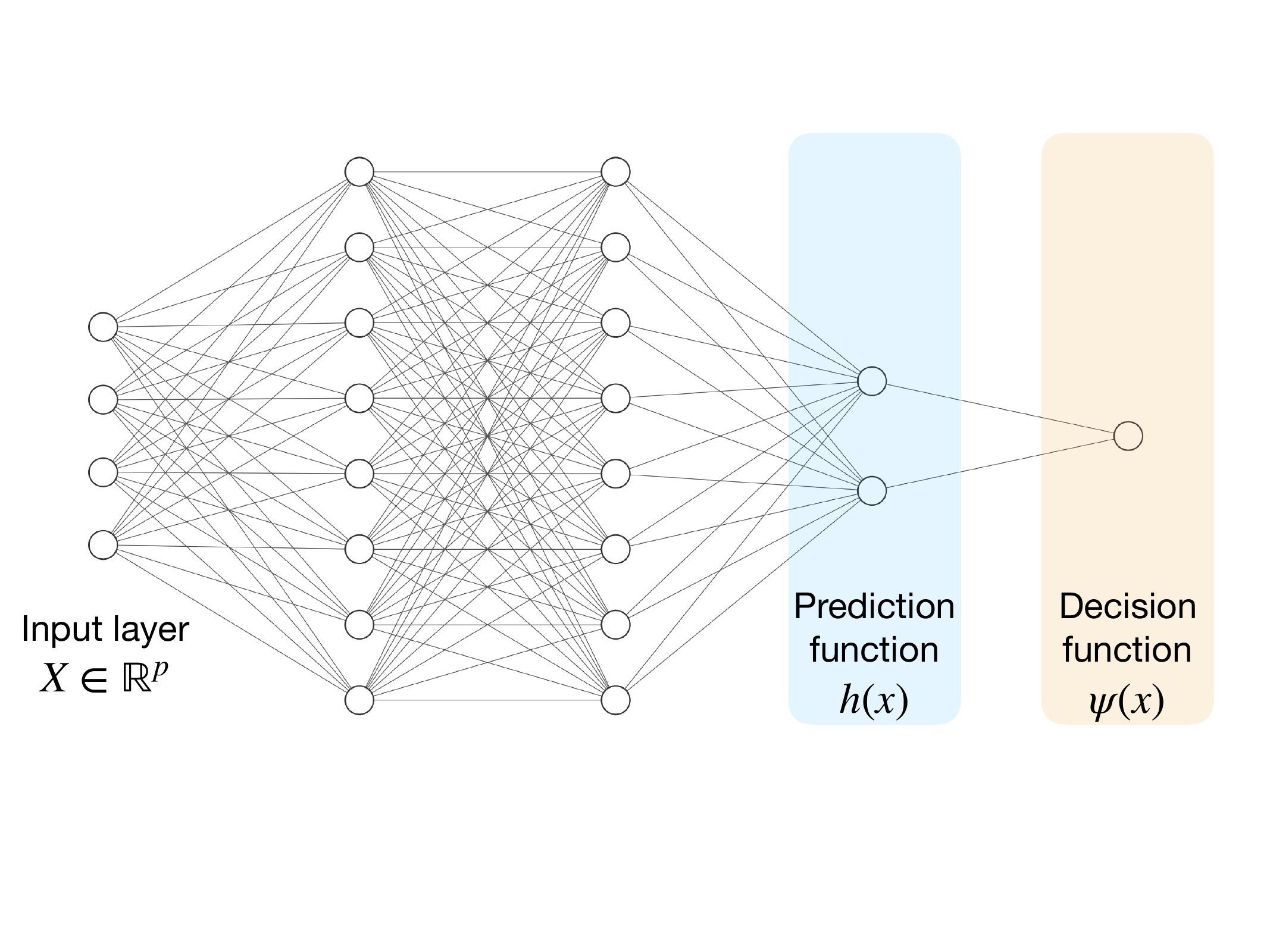}
\caption{
	Example NN architectures for coupling the prediction and decision functions ($h$ and $\psi$, respectively) of a selective prediction model.
	The left architecture couples $h$ and $\psi$ by sharing the last hidden layer of the network.
	The right architecture defines $\psi$ as a function of the prediction function.
	The empirical analyses in this paper define $\psi$ as a logistic function of the entropy (or the expected loss) at $x$, as estimated by the prediction function.
	In this way, the decision function learns to abstain for inputs with high entropy and encourages the prediction network to assign high entropy to outliers and out-of-sample observations.
	This figure appears in color in the electronic version of this article.
}
\label{fig:coupled_models}
\end{figure}

The major steps when implementing pSPS are to decide the model class, select the cost of abstention $\delta$, and solve the optimization problem.
For ease of exposition, we refer to the prediction set function in \eqref{eq:obj_dt} and conditional density function in \eqref{eq:obj_mle} using the single term ``prediction function.''

Our estimation procedure is suitable for any differentiable model class for the decision and prediction functions.
Nevertheless, we recommend selecting a decision model class with higher model complexity than the prediction model class.
For instance, training a simple, highly-interpretable prediction model in conjunction with a nonparametric decision function helps protect against misspecification of the prediction model class.
When there is sufficient data for training more complex prediction models, both the decision and prediction functions can be nonparametric.
We do not recommend fitting parametric decision functions, since the model can over-extrapolate for out-of-distribution samples.

Many empirical analyses in this paper use the NN model class.
Because the decision and prediction functions likely depend on similar latent factors, we couple the decision and prediction function as a single multi-task NN (Figure~\ref{fig:coupled_models}) rather than separate NNs.
This reduces the number of model parameters and, thereby, improves computation speed and performance.
We define this model coupling following results in Theorem~\ref{thrm:population_version}.
Since the population-level decision function thresholds on the optimal expected loss (modulo some scaling terms), we define the decision function as a logistic model of the estimated expected loss.
See Section~\ref{sec:coupling} of the Appendix for more details.

The cost of abstaining, $\delta$, should be defined as the highest acceptable expected loss, per Theorem~\ref{thrm:population_version}.
When taking the MLE approach, this corresponds to the largest acceptable entropy of $Y|X =x$, which is directly proportional to the conditional variance in the case of Gaussian-distributed or binary outcomes.
We follow this entropy-based selection strategy in Sections~\ref{sec:mnist} and \ref{sec:mimic}.
Alternatively, one can fit pSPS models over a grid of $\delta$ values, plot the trade-off between the acceptance rate and the expected loss among accepted observations, and select a $\delta$ that attains a desirable balance between the two operating characteristics \citep{El-Yaniv2010-kd}.
This curve is analogous to how receiver-operating characteristic (ROC) curves are constructed for continuous-valued classifiers.
Various procedures have been proposed for selecting the threshold for continuous-valued classifiers given an ROC curve; we can use analogous procedures for selecting $\delta$.
For example, we can select the inflection point along the curve or the point that achieves a specific acceptance rate or loss.
One may also be able to extend concepts like the Youden Index \citep{Youden1950-eh} to selecting $\delta$.

We minimize the pSPS objective using Adam \citep{Kingma2015-oy} with Monte Carlo sampling to estimate the integral in the uniform acceptance penalty.
That is, we uniformly sample $B$ observations from $\mathcal{X}$ at each iteration to obtain an unbiased estimate of the uniform acceptance penalty and its gradient.
Based on the convergence rates derived in \citet{Kingma2015-oy}, Adam with Monte Carlo sampling is guaranteed to converge for any $B > 0$.
Nevertheless, we recommend choosing a reasonable value for $B$ to reduce the sampling variability.
In this paper, we use $B = 100$ when $p \le 10$ and $B = 2000$ otherwise.
Because the pSPS objective function is non-convex, we recommend running the training procedure with multiple initializations and selecting the one with the smallest training set loss.
We tune penalty parameters $\lambda_0$ and $\lambda_1$ using CV.

\section{Statistical inference for SPS model coverage rates}
\label{sec:calibration}

We now discuss statistical inference for coverage rates of SPS models.
This step is necessary when SPS models are fit using black-box model classes, since they are not guaranteed to achieve the nominal rate \citep{Guo2017-gy,Kuleshov2018-kl}.
The simplest approach is to estimate the marginal coverage rate of a trained SPS model using a held-out labeled dataset.
In this section, we improve on this procedure using $K$-fold CV: we demonstrate that an ensemble of models will have a coverage rate closer to the desired rate of $1 - \alpha$, and we can construct a narrower confidence interval (CI) for this true coverage.
This inference procedure, which we refer to as cross-validation for selection prediction-set models (cSPS), is agnostic to the training procedure.

First, we generalize SPS models, which have been deterministic up to now, to \textit{probabilistic} SPS models.
Such a model is defined by a probabilistic decision function $\hat{\Psi}$, a binary random variable (RV) where 1 is accept and 0 is abstain, and prediction-set function $\hat{H}$, a RV with sample space $S(\mathcal{Y})$.
The marginal coverage rate $\theta$ of the model $(\hat{H}, \hat{\Psi})$ is the rate at which the prediction sets cover the true outcome among the accepted observations, i.e.
\begin{align}
	\theta_{\hat{\Psi}, \hat{H}} = \Pr\left(Y \in \hat{H}(X)  \mid \hat{\Psi}(X) = 1; \hat{\Psi}, \hat{H} \right).
	\label{eq:marginal_cov}
\end{align}
Note that the marginal coverage here is defined for a given model, so the probability in \eqref{eq:marginal_cov} is implicitly conditioned on the training data.

Of course, marginal coverage can differ greatly from the conditional coverage at a particular $x$.
In practice, we cannot evaluate the coverage of a prediction set at a particular $x$ unless there are multiple observations with the same exact feature values.
Instead, we define ``local'' coverage on a subset $\mathcal{A} \subset \mathcal{X} \times \mathcal{Y}$ as
\begin{align}
	\Pr\left(
	Y \in \hat{H}(X)
	\mid
	\hat{\Psi}(X) = 1,
	(X, Y) \in \mathcal{A}
	; \hat{\Psi}, \hat{H}
	\right).
	\label{eq:local_cov}
\end{align}
For example, for binary outcomes, the local coverage for each class is defined using subsets $\mathcal{A}_0 = \{(x, 0): x \in \mathcal{X}\}$ and $\mathcal{A}_1 = \{(x, 1): x \in \mathcal{X}\}$.
Alternatively, we may be interested in the coverage of different subpopulations in $\mathcal{X}$, such as protected groups or minority populations.
In such cases we can perform a more detailed analysis of coverage by evaluating the local coverages in multiple pre-defined subsets.
We focus on calibrating marginal coverage in this paper, but extending the procedure to calibrate local coverage rates is straightforward.

Our results below only require the estimation procedure to converge.
That is, suppose observations are drawn iid from some distribution $P$ and there is a sequence of individual SPS model estimators $(\psi_n, h_n)$ for $n = 1,2,...$ that map $n$ training observations to a particular SPS model.
Denote the estimated SPS model as $(\hat{\psi}_{n}, \hat{h}_{n})$.
We assume that the estimates converge in probability to some SPS model $(\psi, h)$ in $L_2(P)$, i.e.
\begin{align}
\Pr\left(\left \|\hat{\psi}_n - \psi \right \|_{P,2} + \left \|\hat{h}_n - h \right \|_{P,2} \ge \epsilon \right) \rightarrow 0 \quad \forall \epsilon > 0
\label{eq:convergence}
\end{align}
where $\| f \|_{P,2} = (P |f|^2)^{1/2}$ and the probability is taken over the training observations.
For simplicity, suppose the limiting decision function $\psi$ is a binary decision function, though one can consider more general formulations where $\psi$ is probabilistic.
Note that this limiting SPS model does not necessarily have to coincide with the population-level model in Theorem~\ref{thrm:population_version}, which requires a stronger set of assumptions.

\subsection{Inference for the marginal coverage rate using a single data-split}
\label{sec:singel_recalib}

To begin, we describe a simple data-split approach to construct a point estimate and CI for a model's coverage rate.
We partition the data into training and validation sets.
Suppose the validation set $V_n$ has $n_V = n/K$ observations for a fixed integer $K \ge 2$.
(For simplicity, suppose $n$ is a multiple of $K$.)
Given a trained SPS model $(\hat{\psi}_{n - n_V}, \hat{h}_{n - n_V})$ from pSPS, define its probabilistic version where the decision RV $\hat{\Psi}_n(x)$ is equal to one with probability $\hat{\psi}_{n - n_V}(x)$ and the prediction-set RV $\hat{H}_{n}(x)$ is constant with value $\hat{h}_{n - n_V}(x)$.
Let $\theta_{\hat{\Psi}_{n}, \hat{H}_{n}}$ denote the marginal coverage rate of $(\hat{\Psi}_n, \hat{H}_n)$, as defined in \eqref{eq:marginal_cov}.
We estimate its value using $\check{\theta}_{\hat{\Psi}_n, \hat{H}_n, n_V} = \check{\gamma}_{n_V}/\check{q}_{n_V}$ where
$\check{\gamma}_{n_V}
=
\frac{1}{n_V}
\sum_{i \in V_n}
\hat{\psi}_{n - n_V}(x_i)
\mathbbm{1}\left\{
y_i \in \hat{h}_{n - n_V}(x_i)
\right\}$ 
(the empirical coverage on the validation dataset) and $\check{q}_{n_V} = \frac{1}{n_V} \sum_{i \in V_n} \hat{\psi}_{n - n_V}(x_i)$ (the empirical number of accepted observations on the validation dataset).
The asymptotic normal distribution of this coverage estimator is given in the following lemma.
\begin{lemma}
\label{lemma:indiv}
Suppose the observations are sampled iid from some distribution $P$ and the SPS estimates converge as in \eqref{eq:convergence}. Then
\begin{align}
\sqrt{n_V}
\left(
\check{\theta}_{\hat{\Psi}_n, \hat{H}_n, {n_V}} - \theta_{\hat{\Psi}_n, \hat{H}_n}
\right)
\rightarrow_d
N \left(
0,
a_0^\top \Sigma_0 a_0
\right )
\label{eq:indiv_asym}
\end{align}
where
$\Sigma_0 = \Cov\left(\psi(X) \mathbbm{1}\{Y \in h(X) \}, \psi(X) \right )$, $a_0 = (1/q_0, \gamma_0/q_0^2)$, $q_0 = \Pr(\psi(X) = 1)$ and $\gamma_0 = \Pr(\psi(X) \mathbbm{1}\{Y \in \psi(X)\} = 1)$.
The asymptotic variance can be estimated using consistent estimators for $a_0$ and $\Sigma_0$, i.e.
\begin{align}
\hat{a}_{n_V}^\top
\widehat \Cov\left(
\left\{
\left(
\hat{\psi}_n(x_i)
\mathbbm{1}\left\{
y_i \in \hat{h}_n(x_i)
\right\},
\hat{\psi}_n(x_i) \right)
\right\}_{i\in V_n}
\right)
\hat{a}_{n_V}
\label{eq:sigma_single}
\end{align}
where $\hat{a}_{n_V} = \left(
\begin{matrix}
1/\check{q}_{n_V} & -\check{\gamma}_{n_V}/\check{q}_{n_V}^2
\end{matrix}
\right)^\top$.
\end{lemma}
\noindent
The drawback of using a single data-split is that both model estimation and coverage rate estimation are performed using only a portion of the data.
We can instead leverage the dataset fully using $K$-fold CV.

\subsection{Coverage rate for an SPS ensemble using $K$-fold CV}
\label{sec:bag_recalib}
We now describe cSPS, where we train an ensemble of SPS models and perform statistical inference for its marginal coverage using $K$-fold CV.
This ensembling approach uses the entire dataset for both training and statistical inference, which results in better coverage rates and tighter CIs than the single-split method.

The ensemble is constructed as follows.
Given $n$ observations, evenly divide them into $K$ folds, where each fold contains $n_V = n/K$ observations.
For $k = 1,...,K$, let $V_{k,n}$ denote the set of indices in the $k$th fold and $(\hat{\psi}_{k,n - n_V}, \hat{h}_{k,n  - n_V})$ be the individual model trained on all but the $k$th fold.
The ensemble is a uniform mixture of these $K$ models, like that in bagging \citep{Breiman1996-bf}.
Alternatively, we can formulate the ensemble as a \textit{single} probabilistic SPS model $(\hat{\Psi}_n^{\CV}, \hat{H}_n^\CV)$.
At $x$, this model accepts with probability $\frac{1}{K}\sum_{k = 1}^K \hat{\psi}_{k,n - n_V}(x)$ and, upon acceptance, outputs the prediction set $\hat{h}_{k,n - n_V}(x)$ with probability
$\left(
\sum_{i = 1}^K \hat{\psi}_{k, n - n_V}(x)
\right)^{-1}
\hat{\psi}_{k, n - n_V}(x).$

We visualize the ensemble's prediction by overlaying prediction sets from the individual models as shown in Figure~\ref{fig:summary_everything}.
When an individual model chooses to abstain, we display its output as a non-informative prediction set that covers all possible values of the outcome.
For continuous outcomes, this may correspond to the entire real line $[-\infty, \infty]$, unless the outcome is restricted to a more limited set of values.
For categorical outcomes, the set of possible outcomes may include values outside of those observed in the training data.
We visualize this as the set containing all of $\mathcal{Y}$ as well as an ``Anything'' category.

The following result shows that ensembling generally improves coverage rates.
In particular, we prove that the conditional coverage rates of an ensemble are closer to the desired coverage rate of $1 - \alpha$ over the region where all the individual models accept.
In addition, the marginal coverage rate of the ensemble is closer to the desired $1 - \alpha$ rate if the marginal acceptance rates are the same across the individual models.
\begin{theorem}
	Consider SPS models $(\hat{\psi}_k, \hat{h}_k)$ for $k=1,..,K$ where $\hat{\psi}_k: \mathcal{X} \mapsto \{0, 1\}$.
	Let the ensemble of the $K$ models be denoted $(\hat{\Psi}^{\CV}, \hat{H}^{\CV})$.
	Let $A$ be a RV with uniform distribution over $\{1,...,K\}$.
    Let $\tilde{\mathcal{X}} = \cap_{k = 1,..,K} \{x \in \mathcal{X}: \hat{\psi}_{k}(X) = 1\}$.
    Then, the conditional coverage rates of the ensemble are also closer to the desired nominal rate of $1 - \alpha$ than a randomly selected individual model on average, i.e.
    \begin{align}
    \begin{split}
    & \E_X\left[
    \left\{
    \Pr\left(
    Y \in \hat{H}^{\CV}(X) \mid X
    \right)
    - (1 - \alpha)
    \right\}^2
    \mid X \in \tilde{\mathcal{X}}
    \right] \\
    \le &
    E_A\left[
    \E_{X}\left[
    \left\{
    \Pr\left(
    Y \in \hat{h}_{A}(X) \mid X, A
    \right)
    - (1 - \alpha)
    \right\}^2
    \mid X \in \tilde{\mathcal{X}}
    \right]
    \right].
    \end{split}
    \end{align}
    Moreover, if the marginal acceptance probabilities $\Pr(\psi_k(X) = 1)$ are the same for all $k = 1,..,K$, then the marginal coverage rate of the ensemble is closer to the desired rate of $1 - \alpha$ than a randomly selected individual model on average, i.e.
	\begin{align}
	\begin{split}
	\left\{ \Pr\left(
	Y \in \hat{H}^{\CV}(X) \mid \hat{\Psi}^{\CV}(X) = 1
	\right) - (1 - \alpha)\right\}^2 \\
	\le
	E_A\left[
	\left\{ \Pr\left(
	Y \in \hat{h}_{A}(X) \mid \hat{\psi}_{A}(X) = 1, A
	\right) - (1 - \alpha)\right\}^2
	\right].
	\end{split}
	\end{align}
	\label{thrm:bag}
\end{theorem}
\noindent Although the conditions in Theorem~\ref{thrm:bag} may seem overly restrictive, they are increasingly likely to hold as the amount of data increases because the individual decision models converge to the same model $\psi$ in probability (see the proof for Theorem~\ref{thrm:asym_normal} in the Appendix).
Indeed, we observe improved coverage rates for the ensemble in our simulation studies.

We now describe a statistical inference procedure for the coverage of the ensemble model $(\hat{\Psi}_n^\CV, \hat{H}_n^\CV)$.
For the $k$th individual model $(\hat{\psi}_{k,n - n_V}, \hat{h}_{k,n - n_V})$,  let $\check{\gamma}_{k, n_V}$ and $\check{q}_{k, n_V}$ be fold-specific averages defined in Section~\ref{sec:singel_recalib}.
Our estimator for $\theta^\CV_{\hat{\Psi}_n, \hat{H}_n}$ is $\check{\theta}^\CV_{\hat{\Psi}_n, \hat{H}_n, n_V} = \check{\gamma}_{n_V}/\check{q}_{n_V}$ where $\check{\gamma}_{n_V} = \frac{1}{K} \sum_{k=1}^K \check{\gamma}_{k, n_V}$ and $\check{q}_{n_V} = \frac{1}{K} \sum_{k=1}^K \check{q}_{k, n_V}$.
Theorem~\ref{thrm:asym_normal} shows that this estimator is asymptotically normal.
The crux of the proof is to show that the bias introduced by the dependence between the $K$ models is a negligible second-order remainder term.
\begin{theorem}
	\label{thrm:asym_normal}
	Suppose observations are iid samples from some distribution $P$ and the model estimates converge as in \eqref{eq:convergence}.
	Then
	\begin{align}
	\sqrt{n_V}
	\left(
	\check{\theta}^\CV_{\hat{\Psi}_n, \hat{H}_n, n_V} - \theta^{\CV}_{\hat{\Psi}_n, \hat{H}_n}
	\right)
	\rightarrow_d
	N \left(
	0,
	\frac{1}{K}a_0^\top \Sigma_0 a_0
	\right )
	\label{eq:bag_asym}
	\end{align}
	where $a_0$ and $\Sigma_0$ were defined in Lemma~\ref{lemma:indiv}.
	A consistent estimator for the asymptotic variance is $\frac{1}{K^2} \hat{a}_n^\top \hat{\Sigma}_n \hat{a}_n$
	where $\hat{a}_n$ is the $2K \times 1$ vector that tiles the vector
	$
	(\begin{matrix}
	1/\check{q}_{k, n_V} & -\check{\gamma}_{k,n_V}/\check{q}_{k,n_V}^2
	\end{matrix})$
	and $\hat{\Sigma}_n$ is a block diagonal matrix with the $k$th block as
	\begin{align}
		\widehat \Cov\left(
		\left\{
		\left(
		\hat{\psi}_{k,n - n_V}(X_{i})
		\mathbbm{1}\left\{
		Y_{i} \in \hat{h}_{k, n - n_V}(X_{i})
		\right\},
		\hat{\psi}_{k, n - n_V}(X_{i}) \right)
		\right\}_{i \in V_{k,n}}
		\right ).
		\label{eq:cov}
	\end{align}
\end{theorem}
\noindent Comparing Lemma~\ref{lemma:indiv} and Theorem~\ref{thrm:asym_normal}, we find that the asymptotic variance of the ensemble's marginal coverage estimator is $1/K$ of that for the individual model, which means that ensembling reduces the CI width.

We discuss the sensitivity of the ensemble and the statistical inference procedure to the choice of $K$ in Section~\ref{sec:kfolds_choice} of the Appendix, and found minimal differences empirically.
Theoretically, this is because the asymptotic variance of the ensemble is independent of $K$.
Therefore we suggest setting $K$ to commonly recommended values in CV.

Finally, one may extend this statistical inference procedure to perform model selection.
In particular, given a set of candidate SPS models, one may want to select the one whose marginal coverage rate is closest to the desired rate.
Section~\ref{sec:model_select} of the Appendix outlines a simple procedure based on serial gate-keeping \citep{Dmitrienko2007-hp}.

\section{Simulation results}
\label{sec:simulations}

We now evaluate the proposed estimation and inference procedures through simulations.
We simulate $Y|X = x$ to be a normal RV with mean and variance depending on $x$.
In all cases, we use the robust MLE approach to fit decision and density functions and then output estimated PIs.
We estimate the conditional mean and variance using fully-connected NNs with one to three hidden layers.
All models were fit using 1600 training observations unless specified otherwise.
The full simulation settings are given in Section~\ref{sec:sim_settings} in the Appendix, as well as additional simulation results involving higher dimensional datasets.
We note that figures in the electronic version of this article are provided in color, and any mention of color refers to that version.

\begin{figure}
	\begin{tabular}{cp{0.8\linewidth}}
		\raisebox{-.8\height}{\includegraphics[width=0.2\linewidth]{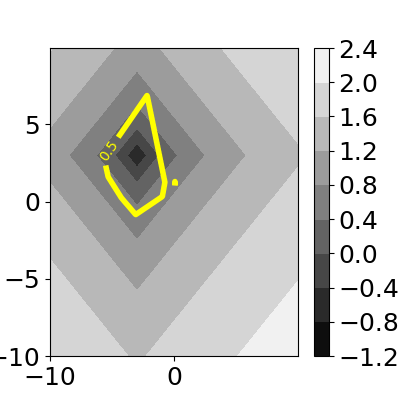}}
		& A) Plot of the decision boundaries for pSPS models fit using the abstention costs $\delta = 0.8, 1.2, 1.6, 2.0$, labeled using the colored lines and entropy of $Y|X = x$, shown using the background shading, over the domain $\mathcal{X} = [-10,10]^2$.
		The colorbar maps entropy to shades of gray.
		The decision boundaries for $\delta$ approximately follow the $\delta$-level-set for entropy.\\
		\midrule
		\raisebox{-.8\height}{\includegraphics[width=0.2\linewidth]{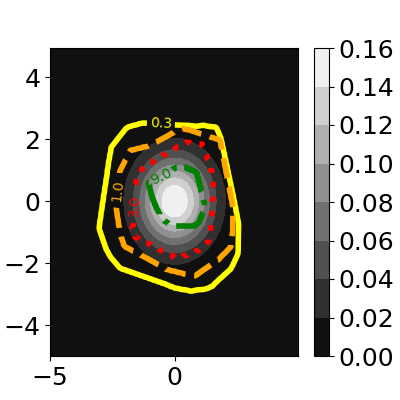}} &
		B) Plot of the decision boundaries for pSPS models fit using penalty parameters $\lambda_1 = 0.3, 1, 3.0, 9.0$, labeled using the colored lines, and the true density of $x$, shown using the background shading, over the domain $\mathcal{X} = [-5,5]^2$.
		The colorbar maps density to the intensity of the black shading.
		The decision boundaries approximately follow the level-sets of density.
		\\
		\midrule
		\raisebox{-.8\height}{\includegraphics[width=0.2\linewidth]{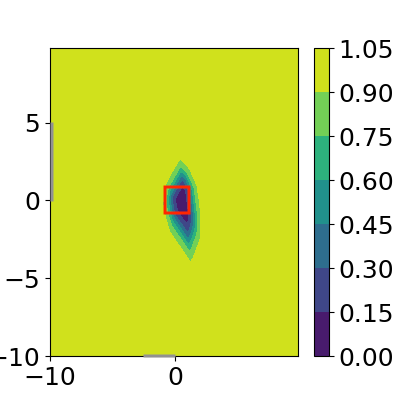}} &
		C) The acceptance probability for a pSPS model over the domain $\mathcal{X} = [-10,10]^2$.
		Here, the true conditional mean of $Y|X=x$ is linear everywhere but over the region $[-1,1]^2$, highlighted by the red box.
		We constrain the pSPS to use a linear function for the prediction model and a neural network for the decision model.
		The resulting model learns to abstain over the misspecified region $[-1,1]^2$.
	\end{tabular}
	\caption{Simulation results for penalized loss minimization for SPS models. This figure appears in color in the electronic version of this article, and any mention of color refers to that version.}
	\label{fig:sims}
\end{figure}

\paragraph{Verifying theoretical results for non-convex settings}
Theorem~\ref{thrm:population_version} implies that (i) if $\lambda_1$ is close to zero, the fitted decision function accepts $x$ if the entropy of $Y|X=x$ is below the cost for declining ($\delta$) and (ii) if the entropy of $Y|X=x$ is constant over $\mathcal{X}$, then the fitted decision function accepts $x$ proportional to the density of $x$.
However, the result assumed that the model was a global minimizer of the risk; this is difficult in practice because the optimization problem is non-convex.
As such, we check the applicability of this result using two simulations.
To test (i), we define $X$ to be a uniform RV over $\mathcal{X} = [-10,10]^2$ and $Y|X = x$ to have variance following a piece-wise linear function in $x$.
We fit pSPS models using $\delta = 0.8, 1.2, 1.6, 2.0$ but fix $\lambda_1 = 0.001$.
The resulting decision functions for each $\delta$ roughly match the $\delta$-level-set for entropy (Fig~\ref{fig:sims}A), which matches the result in Theorem~\ref{thrm:population_version}.
To test (ii), $X$ is generated from a Gaussian distribution with unit variance that is truncated over the domain $\mathcal{X} = [-5, 5]^2$ and $Y|X=x$ has variance one.
We fit pSPS models using $\lambda_1 = 0.3, 1, 3, 9$ but fix $\delta = 2.5$.
In both cases, $\lambda_0$ is fixed at $0.5$.
As seen in Figure~\ref{fig:sims}B, the resulting decision functions now roughly trace the level-sets of $p^*(x)$ instead, again in line with our theoretical results.

\paragraph{Robustness under model misspecification}
One benefit of learning a decision function is protection against model misspecification.
We ran a simulation where $\E[Y|X=x]$ is linear in $x$ over $\mathcal{X} = [-10,10]^2$, except in $[-1,1]^2$ where it follows a quadratic function.
Using pSPS, we fit a linear model for the conditional mean but a multilayer neural network for the decision function.
The decision function learns to accept in all regions but the square in which the linear model is misspecified (Fig~\ref{fig:sims}C).
We note that decision functions that are fit using simple parametric models may not protect against model misspecification.
For an example of such a case, see Figure~\ref{fig:sims_additional} of the Appendix.

\begin{figure}
	\centering
	\includegraphics[width=\linewidth]{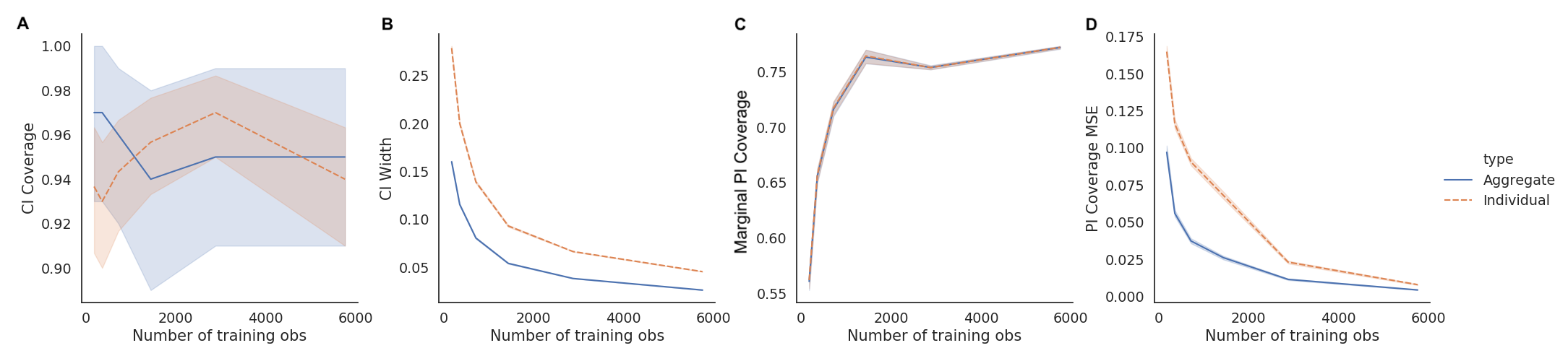}
	\caption{
		We fit pSPS models to estimate 80\% PIs and construct 95\% CI for the marginal coverage rate of the PIs over the accepted observations.
		We fit models via 3-fold CV and compare the performance of the individual models to the ensemble model.
		(A) The 95\%-CI achieves the nominal rate across a range of sample sizes.
		(B) The CIs for the ensemble are consistently narrower than that for individual models.
		(C) The marginal PI coverage of the ensemble and individual models approach the nominal rate of 80\% as sample size increases.
		(D) The mean squared error (MSE) between the conditional PI coverage rate and the nominal rate (for accepted $x$) is smaller for the ensemble.
		This figure appears in color in the electronic version of this article.
	}
	\label{fig:sim_coverage}
\end{figure}

\paragraph{Comparing coverage rates of ensemble and individual models}
Here we compare prediction set coverage rates and their associated CIs for individual and aggregate models from 3-fold CV (Fig.~\ref{fig:sim_coverage}).
We vary the number of training observations from 180 to 5760 and estimate 80\% PIs using pSPS.
The marginal PI coverage rates of the aggregate and individual models are nearly identical at all sample sizes (Fig~\ref{fig:sim_coverage}C), which is unsurprising since the individual models' coverage rates were very similar across folds.
Their marginal coverage rates are initially much lower than the nominal 80\% rate and converge to the nominal rate as the number of observations increases.
Nonetheless, cSPS protects us from being over-optimistic about the PI coverage rates --- the CIs for the marginal PI coverage (with respect to the accepted observations) achieve their nominal 95\% CI coverage rate  (Fig~\ref{fig:sim_coverage}A).
Thus, the CIs reflect the actual PI coverage rates of the black-box SPS model.
Moreover, the width of the CI for the ensemble is consistently half that for individual models (Fig~\ref{fig:sim_coverage}B), as suggested by Theorem~\ref{thrm:asym_normal}.
Finally, we quantify how much the conditional coverage rate for accepted $x$ (i.e. $\Pr(Y \in \hat{h}(X) | X = x)$) differs from the nominal rate in terms of the mean squared error between the two values (Fig~\ref{fig:sim_coverage}D).
In line with our results in Theorem~\ref{thrm:bag}, the conditional coverage rates of the ensemble are closer to the nominal rate, particularly at small sample sizes.

\paragraph{An ablation study}
Finally, we perform an ablation study where we fit selective prediction models for increasingly complex objective functions.
In particular, we consider fitting NNs for selective prediction with no penalties \citep{Bartlett2008-sn}, only the information borrowing penalty \citep{Geifman2019-ve}, and both the information borrowing penalty and the uniform acceptance penalty.
For reference, we also combine an ordinary NN that always accepts with an outlier detector fit using an Isolation Forest \citep{Liu2012-cs}.
As a toy example, we predict a continuous outcome with respect to a single covariate given a training dataset with 100 observations.

\begin{figure}
	\centering
	\includegraphics[width=0.22\linewidth]{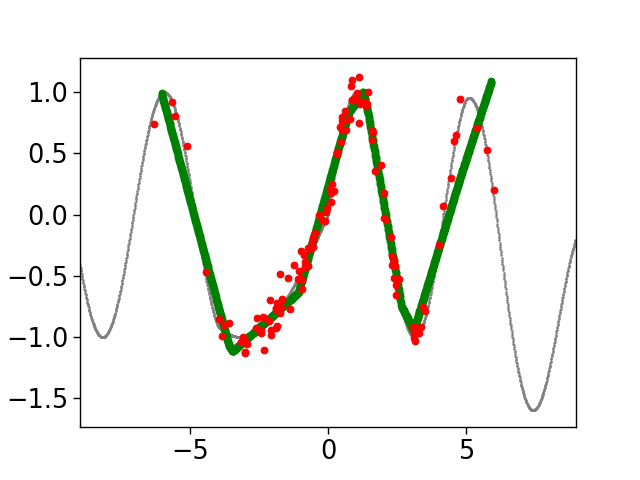}
	\includegraphics[width=0.22\linewidth]{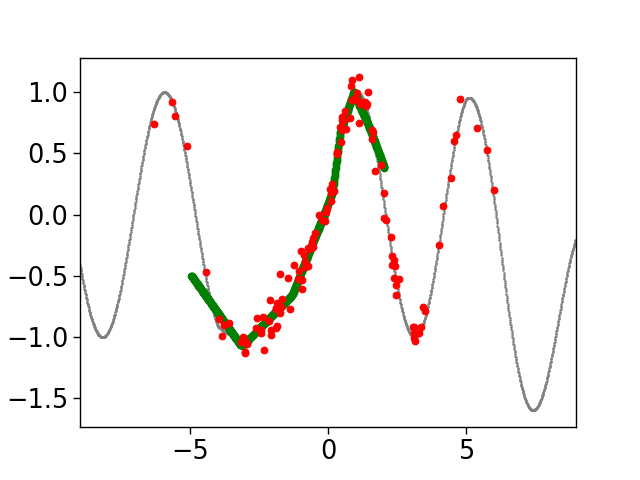}
	\includegraphics[width=0.22\linewidth]{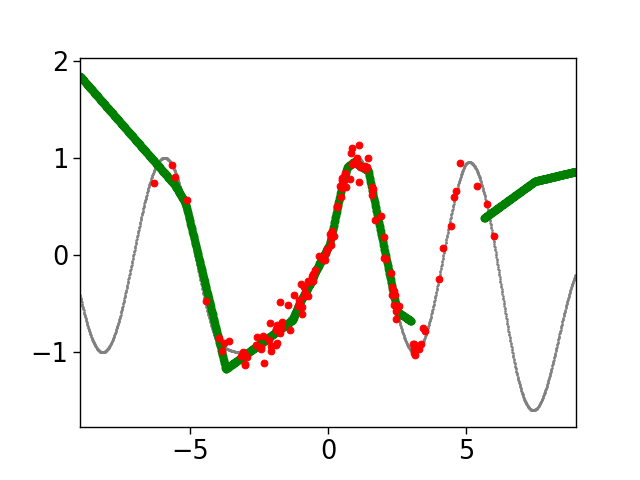}
	\includegraphics[width=0.22\linewidth]{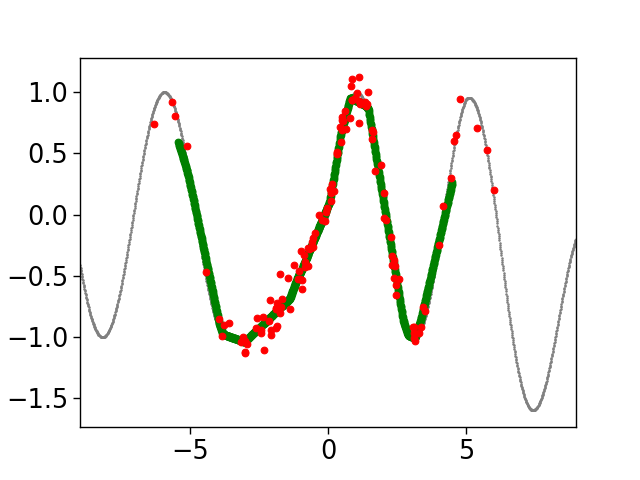}
	\caption{Ablation study on simulated data.
	Fitted models from left to right: accept-all with outlier detector, selective prediction with no penalties, selective prediction with only an information borrowing penalty, selective prediction with both the information borrowing penalty and an uniform acceptance penalty.
	Observed data are displayed as red points. 
	The true model is plotted as the thin gray line. The  predicted conditional mean in the accepted region (defined as acceptance probability $> 0.7$) are plotted by the thick green line.
	This figure appears in color in the electronic version of this article, and any mention of color refers to that version.
	}
	\label{fig:compare}
\end{figure}

We plot the estimated conditional means from the various models in Figure~\ref{fig:compare}.
The accept-all NN combined with an outlier detector was inaccurate in the tails of the distribution.
The selective prediction model fit without any penalties abstained much more often than the other models. Since the objective only depends on accepted observations, the model is trained on less data.
By adding the information borrowing penalty, the model accepted more observations but extrapolated incorrectly beyond the training data.
Finally, by adding the uniform acceptance penalty, the model selected a large region for which it consistently gave accurate predictions.

\section{MNIST digit classification}
\label{sec:mnist}

We now compare pSPS to existing methods in an experiment similar to that in \citet{Lakshminarayanan2017-wn}.
The task is to predict the digits in 28x28 pixel black-and-white images from the MNIST dataset, using a training set with 60,000 images.
To assess how often the models abstain for unseen images, we only train them on images of the digits from zero to eight, but not nine.
We evaluate different SPS models by comparing the test loss of the different methods on the seen digits and acceptance probabilities for images of the unseen digit nine, clothes from the Fashion MNIST dataset, and random noise.
The test loss is defined as the adaptively truncated negative log likelihood over the seen digits with the cost of abstaining as $\delta = 0.3$.
This choice of $\delta$ corresponds to the entropy of a multinomial RV with probability 0.95 of being one class and uniform across the remaining eight classes.
See Section~\ref{sec:mnist_appendix} of the Appendix for additional details.

We compare pSPS to three other approaches, where two of the methods train the decision function separately.
The first trains a pSPS using only the information borrowing penalty (\texttt{Borrow-only pSPS}).
The second fits the conditional density function using MLE and a decision function that thresholds on entropy at $\delta$ (\texttt{Threshold}), which has been suggested in previous literature \citep{Geifman2017-rl} as well as Theorem~\ref{thrm:population_version} in this manuscript.
The third is similar, but the decision function combines entropy-thresholding with an outlier detector (\texttt{Threshold+OD}); here we use the Isolation Forest method \citep{Liu2012-cs}.
We select the threshold of the outlier detector so that the test loss does not change too much.
Since \citet{Lakshminarayanan2017-wn} found that ensembles of NNs are better at identifying regions with high uncertainty compared to individual NNs, all the methods in this section estimate the conditional density function using an ensemble of five NNs.
For the pSPS model, we couple the decision and prediction functions as described in Section~\ref{sec:coupling} of the Appendix.

We consider two ways to model the image data, i) perform dimension reduction by PCA and fit dense NNs with two hidden layers or ii) fit convolutional neural networks (CNNs) on the original 28$\times$28 image data.
The advantage of PCA is its simplicity, since we can easily define a uniform acceptance penalty with respect to the support of the PCA-transformed data. However, CNNs are generally favored for processing image data.

The representation of the image data has important implications on how we define the uniform acceptance penalty.
In particular, the uniform acceptance penalty can only have a meaningful effect on the training procedure if the training data occupies a non-negligible volume in $\mathcal{X}$.
Otherwise, the uniform acceptance penalty will be near-zero and the learned decision boundaries will be too coarse.
When the input variables are the PC scores, we define $\mathcal{X}$ as the smallest bounding box of the PC scores.
Defining the uniform acceptance penalty when the input variable is the raw image requires more care.
For example, we should not define $\mathcal{X}$ to be the set of all 28x28 images.
Instead, we define $\mathcal{X}$ as the inverse PC transform for the smallest bounding box of the PC scores.
Although the inverse images do not look like handwritten digits, the generated data are sufficient for teaching the model to abstain on out-of-distribution images.

pSPS achieves a smaller test loss compared to other methods on in-sample images and has higher abstention rates for out-of-sample images (Table~\ref{table:mnist_log_lik}).
pSPS trained on the raw image data outperformed pSPS trained on only the PC scores.
Since the digits in the training data are easy to predict, all methods accept them with high probability.
The major difference between the methods is the acceptance probabilities for out-of-class images.
If we exclude the uniform acceptance penalty in the pSPS model and only use the information borrowing penalty, the model has a lower acceptance probability for the unseen digit nine, but its acceptance probabilities for Fashion and noise images were very high.
\texttt{Threshold} had similar behavior and assigned high acceptance rates to out-of-class images.
The outlier detector has small acceptance probabilities for Fashion and noise images, but a higher acceptance rate for the digit nine.
This is because the outlier detector fails to discriminate between the digit nine and the other digits.
In contrast, models trained using pSPS have the smallest acceptance probabilities for all categories of out-of-class images.
In summary, pSPS can improve the uncertainty quantification of ensemble NNs.

\begin{table}
	\caption{
	Comparison of methods on MNIST digit classification.
	We compare pSPS to models that abstain by entropy-thresholding (`Threshold') or both entropy-thresholding and an outlier detector (`Threshold+OD').
	The test loss is the adaptively truncated negative log likelihood evaluated over seen digits 0-8.
	Acceptance probabilities are calculated for the unseen digit 9, Fashion MNIST dataset, and white noise.
	Standard error shown in parentheses.
}
\label{table:mnist_log_lik}
	\centering
		\begin{tabular}{l|l|rrrr}
			& & \multicolumn{4}{c}{Acceptance Probability}\\
			& Test loss & Seen digits & Unseen digit & Fashion & Noise \\
			\midrule
			\multicolumn{6}{c}{PCA + dense NN}\\
			pSPS & 0.033 (0.001) & 92.8 (0.1) & 16.4 (0.5) & 7.1 (0.5) & 0.0 (0.0)\\
			Borrow-only pSPS & 0.031 (0.001) & 94.2 (0.1) &21.0 (0.4) &25.3 (1.1) & 24.7 (0.8)\\
			Threshold & 0.034 (0.001) & 94.1 (0.1) & 21.5 (0.4) & 24.4 (1.0) & 24.5 (0.7)\\
			Threshold + OD & 0.051 (0.001) & 93.6 (0.1) & 21.5 (0.4) & 18.3 (0.8)  & 0.0 (0.0)\\
			\midrule
			\multicolumn{6}{c}{Raw image + CNN}\\
			pSPS & 0.027 (0.001) & 94.3 (0.1) & 14.7 (0.6) & 8.8 (0.6) & 0.0 (0.0)\\
			Borrow-only pSPS & 0.027 (0.001) &  96.8 (0.1) & 25.0 (1.0) & 24.8 (0.9) & 22.1 (3.1)\\
			Threshold & 0.033 (0.001) & 96.8 (0.1) & 23.0 (0.6) & 26.0 (0.8) &  15.6 (1.7)\\
			Threshold + OD & 0.043 (0.001) & 96.4 (0.1) & 23.0 (0.6) & 19.3 (0.7) &  0.0 (0.0)
		\end{tabular}
\end{table}

\section{ICU prediction tasks}
\label{sec:mimic}

We now predict in-hospital mortality and length of ICU stay in the MIMIC-III dataset, which contains time series data on physiological measurements collected during ICU admissions for over 40,000 patients \citep{Johnson2016-gu}.
The dataset is composed of neonatals and patients ages 15 to 90.
We compare pSPS to training a prediction model and thresholding on the predicted value, as well as combining the thresholded model with an outlier detector.
See Section~\ref{sec:mimic_appendix} of the Appendix for experimental details.

Because in-hospital mortality is a binary outcome, we used the robust MLE approach and output 90\% prediction sets based on the estimated model.
We set the cost $\delta$ to 0.5, which is the entropy of a binary RV with success probability $> 0.8$ or $ < 0.2$.
We estimate the 80\% PI for the LOS (in log hours) using the decision-theoretic approach with the absolute discrepancy loss.
We set $\delta = 0.25$, which is the loss of a 80\% PI of the form $[l, 3 \times l]$ and the true outcome, when uncovered, is on average a factor of 2 away from the interval boundaries.

We trained densely-connected NNs with two hidden layers for all prediction models.
For pSPS models, we coupled the decision and prediction models, where the decision model depended on the predicted entropy for the in-hospital mortality task and PI width for the LOS prediction task.
We transformed the continuous covariates using PCA to define $\mathcal{X}$ to be as small as possible.
The resulting data had 122 variables, in which 120 were PC scores and 2 were discrete variables.

We first compared the prediction accuracy of the methods for patients over and under fifty years old, where the models only train on the former age group (Table~\ref{table:mimic_loss_comparison_methods}).
To prevent the methods from directly thresholding on the age variable, we removed the age covariate from the dataset (though there may be other covariates that are correlated with age).
This introduces an interesting challenge, as the definition of an ``out-of-sample'' patient is no longer clear.
For example, a patient who is 51 years old is likely quite similar to one who is 49 years old.
Thus, our goal is not necessarily to simply abstain for the younger patients.
Rather, we want a model that achieves low test loss.

\begin{table}
\caption{
	Performance on in-hospital mortality (top) and LOS (bottom) prediction tasks for the MIMIC dataset.
	We compare pSPS to thresholding on entropy/interval length (`Threshold') or thresholding and using an outlier detector (`Threshold+OD').
	Models are trained on patients $\ge 50$ years old.
	Test loss is the adaptively truncated negative log likelihood (top) and absolute discrepancy (bottom).
	Standard errors shown in parentheses.
}
	\centering
		\begin{tabular}{l|llll}
			Method & Test loss $\le50$yr & Accept $\le 50$yr & Test loss $>50$yr & Accept $>50$yr \\
			\toprule
			\multicolumn{5}{c}{\textit{In-hospital Mortality}}\\
			pSPS & 0.478 (0.032) & 42.38 (3.65) & 0.3376 (0.006) & 80.86 (0.16)\\
			Borrow-only pSPS & 0.689 (0.059) &54.67 (3.06)  & 0.3376 (0.008) & 81.10 (0.22) \\
			Threshold & 0.952 (0.073) & 66.19 (3.59) & 0.3329 (0.004) & 79.59 (0.37) \\
			Threshold+OD & 0.768 (0.047) & 51.44 (2.44) & 0.3768 (0.006) & 73.71 (0.33)\\
			\toprule
			\multicolumn{5}{c}{\textit{Length of Stay}}\\
			pSPS & 0.315 (0.005) & 69.38 (0.55) & 0.2163 (0.002) & 75.41 (0.34) \\
			Borrow-only pSPS & 0.364 (0.010) & 94.37 (0.62) & 0.2078 (0.002) & 97.55 (0.10) \\
			Threshold & 0.367 (0.006) & 98.95 (0.30) & 0.2079 (0.003) & 99.48 (0.07) \\
			Threshold+OD & 0.342 (0.004) & 78.96 (0.08) & 0.2119 (0.003) & 90.45 (0.03) \\
		\end{tabular}

\label{table:mimic_loss_comparison_methods}
\end{table}

In both prediction tasks, pSPS models outperformed the other methods on both patients under and over fifty  (Table~\ref{table:mimic_loss_comparison_methods}).
For the in-hospital mortality task, the acceptance rates and the test loss were similar across the different methods in the older age group.
The pSPS model learned to abstain more for the younger age group than all other methods, resulting in a smaller test loss.
For the LOS task, the pSPS model learned to abstain more often in both age groups.
Its test loss was similar to the other methods in the older age group and was lower in the younger age group.
Thus, pSPS learned to be more conservative when extrapolating in the younger age groups, resulting in better predictions overall.

Next, we fit pSPS models using 5-fold sample-splits over patients from all age groups.
We hold out 6000 randomly selected patients to evaluate marginal coverage rates and compare against estimates from using a single data-split and 5-fold CV (Table~\ref{table:mimic_coverage}).
The constructed CIs cover the test coverage in all but one case.
The marginal coverage rate of the ensemble is between those of the individual models and close to their desired nominal rates.
Its CI is roughly half the width of that for individual models.

\begin{figure}
\centering
\includegraphics[width=0.4\linewidth]{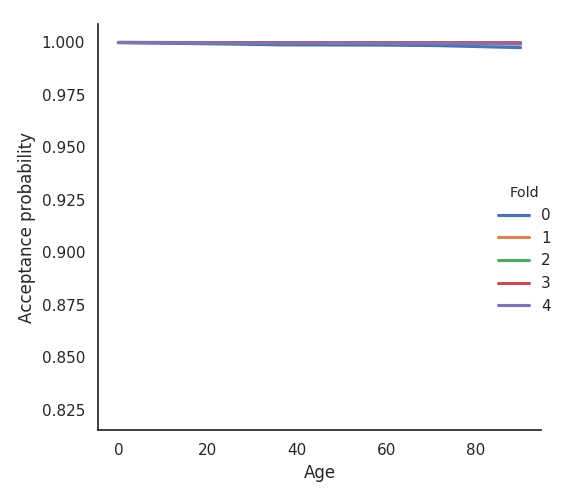}
\includegraphics[width=0.4\linewidth]{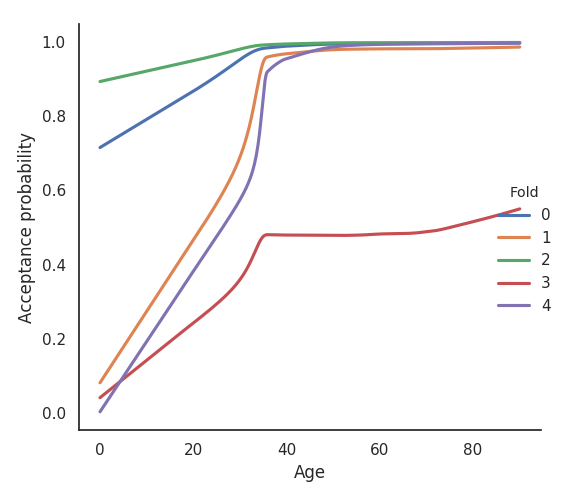}
\caption{Acceptance probabilities for pSPS models fit on patients from all age groups in the MIMIC dataset.
We show how acceptance rates vary with age in the in-hospital mortality (left) and LOS (right) prediction tasks.
Acceptance rates are shown for the individual models fit using 5-fold CV.
This figure appears in color in the electronic version of this article.
}
\label{fig:accept_v_age}
\end{figure}

Figure~\ref{fig:accept_v_age} displays how acceptance probabilities vary with age for the final models fit on all age groups.
Having trained on more data, the final in-hospital mortality SPS model assigned high acceptance probability across all ages.
In contrast, the final LOS model assigned much lower acceptance probabilities to younger patients, because  LOS in the neonatal ICU was highly variable.
See Figure~\ref{fig:example_mimic_predictions} in the Appendix for example predictions from the final model.

\begin{table}
	\caption{
		PI coverage estimates for accepted observations in individual and ensemble SPS models fit on the MIMIC dataset using 5-fold CV.
		The PI coverage on test set is shown for comparison.
	}
	\label{table:mimic_coverage}
	\centering
	\small{
		\begin{tabular}{l|cc|cc}
			& \multicolumn{2}{c}{\textit{In-hospital Mortality}} & \multicolumn{2}{|c}{\textit{Length of Stay}}\\
			& PI coverage est (95\% CI) & Test PI coverage & PI coverage est (95\% CI) & Test PI coverage\\
			\toprule
			Fold 0 & 0.928 (0.918, 0.938) & 0.932 & 0.736 (0.720, 0.753) & 0.740 \\
			Fold 1 & 0.946 (0.938, 0.955) &0.941 & 0.759 (0.743, 0.775) &0.767 \\
			Fold 2 & 0.928 (0.918, 0.937) & 0.932 & 0.710 (0.693, 0.728) & 0.706 \\
			Fold 3 & 0.944 (0.935, 0.952) &  0.949 & 0.798 (0.783, 0.812) & 0.786\\
			Fold 4 & 0.945 (0.936, 0.953) & 0.943 & 0.785 (0.770, 0.799) & 0.775\\
			Ensemble & 0.938 (0.934, 0.942) &  0.939 & 0.759 (0.752, 0.766) & 0.756
		\end{tabular}
	}
\end{table}

\section{Discussion}

We have presented a new penalized empirical loss framework for training SPS models.
The pSPS framework can train existing ML models to learn the limits of their predictive capabilities and abstain when they are uncertain.
Although we have primarily focused on training SPS models using NNs, one can also use other semi- and non-parametric models like random forests and smoothing splines; the training procedures would need to be adapted to optimize the pSPS objective instead.
We then show how to use $K$-fold CV to train ensembled SPS models and perform statistical inference for its marginal coverage rate over the accepted observations.
The resulting model enjoys many desirable properties: it has trained on all the data, its conditional coverage rates are closer to the nominal level, and the CI for its marginal coverage rate is narrower.

One direction of future work is to extend the uniform acceptance penalty to a more targeted version, by utilizing prior knowledge of out-of-distribution samples.
In addition, this work only achieves uniform conditional coverage asymptotically; future studies should explore how to maximize uniformity of the conditional coverage rates in finite sample sizes.

\section*{Data Availability Statement}
The Medical Information Mart for Intensive Care III (MIMIC-III) and Modified National Institute of Standards and Technology (MNIST) datasets used in this paper are openly available at \url{https://doi.org/10.13026/C2XW26} \citep{Johnson2016-gu} and \url{https://pytorch.org/vision/stable/datasets.html#mnist} \citep{deng2012mnist}, respectively.

\bibliographystyle{plain}
\bibliography{main}

\begin{thebibliography}{37}
\providecommand{\natexlab}[1]{#1}
\providecommand{\url}[1]{\texttt{#1}}
\expandafter\ifx\csname urlstyle\endcsname\relax
  \providecommand{\doi}[1]{doi: #1}\else
  \providecommand{\doi}{doi: \begingroup \urlstyle{rm}\Url}\fi

\bibitem[Abadi et~al.(2016)Abadi, Agarwal, Barham, Brevdo, Chen, Citro,
  Corrado, Davis, Dean, Devin, Ghemawat, Goodfellow, Harp, Irving, Isard, Jia,
  Jozefowicz, Kaiser, Kudlur, Levenberg, Mane, Monga, Moore, Murray, Olah,
  Schuster, Shlens, Steiner, Sutskever, Talwar, Tucker, Vanhoucke, Vasudevan,
  Viegas, Vinyals, Warden, Wattenberg, Wicke, Yu, and Zheng]{Abadi2016-zh}
M.~Abadi, A.~Agarwal, P.~Barham, E.~Brevdo, Z.~Chen, C.~Citro, G.~S. Corrado,
  A.~Davis, J.~Dean, M.~Devin, S.~Ghemawat, I.~Goodfellow, A.~Harp, G.~Irving,
  M.~Isard, Y.~Jia, R.~Jozefowicz, L.~Kaiser, M.~Kudlur, J.~Levenberg, D.~Mane,
  R.~Monga, S.~Moore, D.~Murray, C.~Olah, M.~Schuster, J.~Shlens, B.~Steiner,
  I.~Sutskever, K.~Talwar, P.~Tucker, V.~Vanhoucke, V.~Vasudevan, F.~Viegas,
  O.~Vinyals, P.~Warden, M.~Wattenberg, M.~Wicke, Y.~Yu, and X.~Zheng.
\newblock {TensorFlow}: {Large-Scale} machine learning on heterogeneous
  distributed systems.
\newblock 2016.
\newblock URL \url{http://arxiv.org/abs/1603.04467}.

\bibitem[Barber et~al.(2019)Barber, Cand{\`e}s, Ramdas, and
  Tibshirani]{Barber2019-fr}
R.~F. Barber, E.~J. Cand{\`e}s, A.~Ramdas, and R.~J. Tibshirani.
\newblock The limits of distribution-free conditional predictive inference.
\newblock 2019.

\bibitem[Bartlett and Wegkamp(2008)]{Bartlett2008-sn}
P.~L. Bartlett and M.~H. Wegkamp.
\newblock Classification with a reject option using a hinge loss.
\newblock \emph{J. Mach. Learn. Res.}, 9\penalty0 (Aug):\penalty0 1823--1840,
  2008.
\newblock ISSN 1532-4435, 1533-7928.

\bibitem[Benjamens et~al.(2020)Benjamens, Dhunnoo, and
  Mesk{\'o}]{Benjamens2020-eh}
S.~Benjamens, P.~Dhunnoo, and B.~Mesk{\'o}.
\newblock The state of artificial intelligence-based {FDA-approved} medical
  devices and algorithms: an online database.
\newblock \emph{NPJ Digit Med}, 3:\penalty0 118, 2020.
\newblock URL \url{http://dx.doi.org/10.1038/s41746-020-00324-0}.

\bibitem[Breiman(1996)]{Breiman1996-bf}
L.~Breiman.
\newblock Bagging predictors.
\newblock \emph{Mach. Learn.}, 24\penalty0 (2):\penalty0 123--140, 1996.
\newblock ISSN 0885-6125, 1573-0565.
\newblock \doi{10.1023/A:1018054314350}.

\bibitem[Challen et~al.(2019)Challen, Denny, Pitt, Gompels, Edwards, and
  Tsaneva-Atanasova]{Challen2019-zn}
R.~Challen, J.~Denny, M.~Pitt, L.~Gompels, T.~Edwards, and
  K.~Tsaneva-Atanasova.
\newblock Artificial intelligence, bias and clinical safety.
\newblock \emph{BMJ Qual. Saf.}, 28\penalty0 (3):\penalty0 231--237, 2019.
\newblock URL \url{http://dx.doi.org/10.1136/bmjqs-2018-008370}.

\bibitem[Chow(1970)]{Chow1970-uc}
C.~Chow.
\newblock On optimum recognition error and reject tradeoff.
\newblock \emph{IEEE Trans. Inf. Theory}, 16\penalty0 (1):\penalty0 41--46,
  1970.
\newblock ISSN 0018-9448.
\newblock \doi{10.1109/TIT.1970.1054406}.

\bibitem[Deng(2012)]{deng2012mnist}
L.~Deng.
\newblock The mnist database of handwritten digit images for machine learning
  research.
\newblock \emph{IEEE Signal Processing Magazine}, 29\penalty0 (6):\penalty0
  141--142, 2012.

\bibitem[Dmitrienko and Tamhane(2007)]{Dmitrienko2007-hp}
A.~Dmitrienko and A.~C. Tamhane.
\newblock Gatekeeping procedures with clinical trial applications.
\newblock \emph{Pharm. Stat.}, 6\penalty0 (3):\penalty0 171--180, 2007.
\newblock ISSN 1539-1604.
\newblock \doi{10.1002/pst.291}.

\bibitem[El-Yaniv and Wiener(2010)]{El-Yaniv2010-kd}
R.~El-Yaniv and Y.~Wiener.
\newblock On the foundations of noise-free selective classification.
\newblock \emph{J. Mach. Learn. Res.}, 11\penalty0 (May):\penalty0 1605--1641,
  2010.
\newblock ISSN 1532-4435, 1533-7928.

\bibitem[Ettema et~al.(2010)Ettema, Peelen, Schuurmans, Nierich, Kalkman, and
  Moons]{ettema2010prediction}
R.~G. Ettema, L.~M. Peelen, M.~J. Schuurmans, A.~P. Nierich, C.~J. Kalkman, and
  K.~G. Moons.
\newblock Prediction models for prolonged intensive care unit stay after
  cardiac surgery: systematic review and validation study.
\newblock \emph{Circulation}, 122\penalty0 (7):\penalty0 682--689, 2010.

\bibitem[Geifman and El-Yaniv(2017)]{Geifman2017-rl}
Y.~Geifman and R.~El-Yaniv.
\newblock Selective classification for deep neural networks.
\newblock In \emph{Advances in Neural Information Processing Systems 30}, pages
  4878--4887. Curran Associates, Inc., 2017.

\bibitem[Geifman and El-Yaniv(2019)]{Geifman2019-ve}
Y.~Geifman and R.~El-Yaniv.
\newblock {{S}elective{N}et}: A deep neural network with an integrated reject
  option.
\newblock In \emph{Proceedings of the 36th International Conference on Machine
  Learning}, volume~97 of \emph{Proceedings of Machine Learning Research},
  pages 2151--2159. PMLR, 2019.
\newblock URL \url{http://proceedings.mlr.press/v97/geifman19a.html}.

\bibitem[Guo et~al.(2017)Guo, Pleiss, Sun, and Weinberger]{Guo2017-gy}
C.~Guo, G.~Pleiss, Y.~Sun, and K.~Q. Weinberger.
\newblock On calibration of modern neural networks.
\newblock \emph{International Conference on Machine Learning}, 70:\penalty0
  1321--1330, 2017.
\newblock URL \url{http://proceedings.mlr.press/v70/guo17a.html}.

\bibitem[Harutyunyan et~al.(2017)Harutyunyan, Khachatrian, Kale, Ver~Steeg, and
  Galstyan]{Harutyunyan2017-hw}
H.~Harutyunyan, H.~Khachatrian, D.~C. Kale, G.~Ver~Steeg, and A.~Galstyan.
\newblock Multitask learning and benchmarking with clinical time series data.
\newblock 2017.

\bibitem[Johnson et~al.(2016)Johnson, Pollard, Shen, Lehman, Feng, Ghassemi,
  Moody, Szolovits, Celi, and Mark]{Johnson2016-gu}
A.~E.~W. Johnson, T.~J. Pollard, L.~Shen, L.-W.~H. Lehman, M.~Feng,
  M.~Ghassemi, B.~Moody, P.~Szolovits, L.~A. Celi, and R.~G. Mark.
\newblock {MIMIC-III}, a freely accessible critical care database.
\newblock \emph{Sci Data}, 3:\penalty0 160035, 2016.
\newblock ISSN 2052-4463.
\newblock \doi{10.1038/sdata.2016.35}.

\bibitem[Kingma and Ba(2015)]{Kingma2015-oy}
D.~P. Kingma and J.~Ba.
\newblock Adam: A method for stochastic optimization.
\newblock \emph{International Conference for Learning Representations}, 2015.
\newblock URL \url{http://arxiv.org/abs/1412.6980}.

\bibitem[Kosel and Heagerty(2019)]{Kosel2019-ji}
A.~E. Kosel and P.~J. Heagerty.
\newblock Semi-supervised neighborhoods and localized patient outcome
  prediction.
\newblock \emph{Biostatistics}, 20\penalty0 (3):\penalty0 517--541, 2019.
\newblock ISSN 1465-4644, 1468-4357.
\newblock \doi{10.1093/biostatistics/kxy015}.

\bibitem[Kuleshov and Liang(2015)]{Kuleshov2015-yg}
V.~Kuleshov and P.~S. Liang.
\newblock Calibrated structured prediction.
\newblock In \emph{Advances in Neural Information Processing Systems 28}, pages
  3474--3482. Curran Associates, Inc., 2015.

\bibitem[Kuleshov et~al.(2018)Kuleshov, Fenner, and Ermon]{Kuleshov2018-kl}
V.~Kuleshov, N.~Fenner, and S.~Ermon.
\newblock Accurate uncertainties for deep learning using calibrated regression.
\newblock \emph{International Conference on Machine Learning}, 80:\penalty0
  2796--2804, 2018.

\bibitem[Lakshminarayanan et~al.(2017)Lakshminarayanan, Pritzel, and
  Blundell]{Lakshminarayanan2017-wn}
B.~Lakshminarayanan, A.~Pritzel, and C.~Blundell.
\newblock Simple and scalable predictive uncertainty estimation using deep
  ensembles.
\newblock In \emph{Advances in Neural Information Processing Systems 30}, pages
  6402--6413. Curran Associates, Inc., 2017.

\bibitem[Lei and Wasserman(2014)]{Lei2014-uh}
J.~Lei and L.~Wasserman.
\newblock Distribution‐free prediction bands for non‐parametric regression.
\newblock \emph{Journal of the Royal Statistical Society, Series B},
  76\penalty0 (1), 2014.

\bibitem[Liu et~al.(2012)Liu, Ting, and Zhou]{Liu2012-cs}
F.~T. Liu, K.~M. Ting, and Z.-H. Zhou.
\newblock {Isolation-Based} anomaly detection.
\newblock \emph{ACM Trans. Knowl. Discov. Data}, 6\penalty0 (1):\penalty0
  3:1--3:39, 2012.
\newblock ISSN 1556-4681.
\newblock \doi{10.1145/2133360.2133363}.

\bibitem[Muehlematter et~al.(2021)Muehlematter, Daniore, and
  Vokinger]{Muehlematter2021-iq}
U.~J. Muehlematter, P.~Daniore, and K.~N. Vokinger.
\newblock Approval of artificial intelligence and machine learning-based
  medical devices in the {USA} and europe (2015-20): a comparative analysis.
\newblock \emph{Lancet Digit Health}, 3\penalty0 (3):\penalty0 e195--e203,
  2021.
\newblock URL \url{http://dx.doi.org/10.1016/S2589-7500(20)30292-2}.

\bibitem[{Office of the Commissioner}(2018)]{Office_of_the_Commissioner2018-aa}
{Office of the Commissioner}.
\newblock {FDA} permits marketing of artificial intelligence-based device to
  detect certain diabetes-related eye problems.
\newblock
  \url{https://www.fda.gov/news-events/press-announcements/fda-permits-marketing-artificial-intelligence-based-device-detect-certain-diabetes-related-eye},
  2018.
\newblock URL
  \url{https://www.fda.gov/news-events/press-announcements/fda-permits-marketing-artificial-intelligence-based-device-detect-certain-diabetes-related-eye}.
\newblock Accessed: 2021-10-2.

\bibitem[Pimentel et~al.(2014)Pimentel, Clifton, Clifton, and
  Tarassenko]{Pimentel2014-uw}
M.~A.~F. Pimentel, D.~A. Clifton, L.~Clifton, and L.~Tarassenko.
\newblock A review of novelty detection.
\newblock \emph{Signal Processing}, 99:\penalty0 215--249, 2014.
\newblock ISSN 0165-1684.
\newblock \doi{10.1016/j.sigpro.2013.12.026}.

\bibitem[Platt(1999)]{Platt1999-yw}
J.~C. Platt.
\newblock Probabilistic outputs for support vector machines and comparisons to
  regularized likelihood methods.
\newblock In \emph{{ADVANCES} {IN} {LARGE} {MARGIN} {CLASSIFIERS}}, 1999.

\bibitem[Rice et~al.(2008)Rice, Lumley, and Szpiro]{Rice2008-ia}
K.~M. Rice, T.~Lumley, and A.~A. Szpiro.
\newblock Trading bias for precision: Decision theory for intervals and sets.
\newblock \emph{UW Biostatistics Working Paper Series}, 2008.

\bibitem[Robinson et~al.(1966)Robinson, Davis, and Leifer]{Robinson1966-wu}
G.~H. Robinson, L.~E. Davis, and R.~P. Leifer.
\newblock Prediction of hospital length of stay.
\newblock \emph{Health Serv. Res.}, 1\penalty0 (3):\penalty0 287--300, 1966.
\newblock URL \url{https://www.ncbi.nlm.nih.gov/pubmed/5971638}.

\bibitem[Sadinle et~al.(2019)Sadinle, Lei, and Wasserman]{Sadinle2019-ph}
M.~Sadinle, J.~Lei, and L.~Wasserman.
\newblock Least ambiguous set-valued classifiers with bounded error levels.
\newblock \emph{J. Am. Stat. Assoc.}, 2019.
\newblock URL
  \url{https://www.tandfonline.com/doi/abs/10.1080/01621459.2017.1395341}.

\bibitem[Taylor(2000)]{Taylor2000-iu}
J.~W. Taylor.
\newblock A quantile regression neural network approach to estimating the
  conditional density of multiperiod returns.
\newblock \emph{J. Forecast.}, 19\penalty0 (4):\penalty0 299--311, 2000.
\newblock ISSN 0277-6693.

\bibitem[Tortorella(2000)]{Tortorella2000-yc}
F.~Tortorella.
\newblock An optimal reject rule for binary classifiers.
\newblock In \emph{Advances in Pattern Recognition}, pages 611--620. Springer
  Berlin Heidelberg, 2000.
\newblock \doi{10.1007/3-540-44522-6\_63}.

\bibitem[Verburg et~al.(2017)Verburg, Atashi, Eslami, Holman, Abu-Hanna,
  de~Jonge, Peek, and de~Keizer]{verburg2017models}
I.~W.~M. Verburg, A.~Atashi, S.~Eslami, R.~Holman, A.~Abu-Hanna, E.~de~Jonge,
  N.~Peek, and N.~F. de~Keizer.
\newblock Which models can {I} use to predict adult {ICU} length of stay? a
  systematic review.
\newblock \emph{Critical care medicine}, 45\penalty0 (2):\penalty0 e222--e231,
  2017.

\bibitem[Vovk et~al.(2005)Vovk, Gammerman, and Shafer]{Vovk2005-wf}
V.~Vovk, A.~Gammerman, and G.~Shafer.
\newblock \emph{Algorithmic Learning in a Random World}.
\newblock Springer, Boston, MA, 2005.
\newblock URL \url{https://link.springer.com/book/10.1007/b106715}.

\bibitem[Youden(1950)]{Youden1950-eh}
W.~J. Youden.
\newblock Index for rating diagnostic tests.
\newblock \emph{Cancer}, 3\penalty0 (1):\penalty0 32--35, 1950.
\newblock URL
  \url{http://dx.doi.org/10.1002/1097-0142(1950)3:1<32::aid-cncr2820030106>3.0.co;2-3}.

\bibitem[Zadrozny and Elkan(2002)]{Zadrozny2002-ro}
B.~Zadrozny and C.~Elkan.
\newblock Transforming classifier scores into accurate multiclass probability
  estimates.
\newblock In \emph{Proceedings of the Eighth {ACM} {SIGKDD} International
  Conference on Knowledge Discovery and Data Mining}, KDD '02, pages 694--699,
  New York, NY, USA, 2002. ACM.
\newblock ISBN 9781581135671.
\newblock \doi{10.1145/775047.775151}.

\bibitem[Zimmerman et~al.(2006)Zimmerman, Kramer, McNair, Malila, and
  Shaffer]{zimmerman2006intensive}
J.~E. Zimmerman, A.~A. Kramer, D.~S. McNair, F.~M. Malila, and V.~L. Shaffer.
\newblock Intensive care unit length of stay: Benchmarking based on acute
  physiology and chronic health evaluation ({APACHE}) {IV}.
\newblock \emph{Critical care medicine}, 34\penalty0 (10):\penalty0 2517--2529,
  2006.

\end{thebibliography}

\section*{Supporting Information}
The Web Appendix, Figures, and Tables referenced in Sections~\ref{sec:dec-pred}, \ref{sec:calibration}, \ref{sec:simulations}, \ref{sec:mnist}, and \ref{sec:mimic} are available with this paper at the Biometrics website on Wiley Online Library.
Code for reproducing the paper results and fitting selective prediction-set models is available both with this paper on Wiley Online Library and at \url{http://github.com/jjfeng/pc_SPS}.

\appendix
\section{Penalized estimation for SPS models}

\begin{proof}[Proof for Theorem~\ref{thrm:population_version}]
	We will only show the proof for the population-level models for the decision-theoretic approach, since the proof for the robust maximum log likelihood approach is very similar.
	At every $x \in \mathcal{X}$, the population-level estimate $\psi_{\alpha, \delta, \lambda_0, \lambda_1}(x), h_{\alpha, \delta, \lambda_0, \lambda_1}(x)$ must be a solution to
	\begin{align}
		\begin{split}
			\min_{\tilde{\psi} \in [0,1], \tilde{h}  \in \mathcal{S}(\mathcal{Y})}
			& \left( \left( E\left[\ell_{\alpha}\left( \tilde{h}, Y \right) \mid X = x \right]  - \delta \right) p^*(x) +\lambda_1 \right) \tilde{\psi}
			+ \lambda_0 E\left[\ell_{\alpha}\left( \tilde{h}, Y \right) \mid X = x \right] p^*(x).
		\end{split}
		\label{eq:integrand}
	\end{align}
	Minimizing \eqref{eq:integrand} with respect to only $\tilde{h}$ and keeping $\tilde{\psi}$ fixed, the solution at any $x$ such that $p^*(x) > 0$ is
	\begin{align}
		h_{\alpha, \delta, \lambda_0, \lambda_1}(x) \in \argmin_{\tilde{h} \in \mathcal{S}} E\left[\ell_{\alpha}\left( \tilde{h}, Y \right) \mid X = x \right].
	\end{align}
	Let $\ell^*_\alpha(x) = E\left[\ell_{\alpha}\left( h_{\alpha, \delta, \lambda_0, \lambda_1}(x), Y \right) \mid X = x \right]$.
	Then $\psi_{\alpha, \delta, \lambda_0, \lambda_1}(x)(x)$ is 1 if $\left(\ell_{\alpha}^*(x) - \delta\right)p^*(x) + \lambda_1 < 0$ and 0 if $\left(\ell_{\alpha}^*(x) - \delta\right)p^*(x) + \lambda_1 > 0$.
	The behavior is indeterminate otherwise.
\end{proof}

\subsection{Connection between the two approaches}

The decision theoretic and the robust maximum likelihood estimation approach estimate the same population-level selective prediction-set model under the following conditions.
We omit the proof as it is straightforward.
\begin{theorem}
	Suppose all assumptions in Theorem~\ref{thrm:population_version} hold.
	Let $\mathcal{X}' = \{x: x\in \mathcal{X}, p^*(x) > 0\}$.
	Let $\delta, \alpha, \lambda_0, \lambda_1 \ge 0$ and $\mathcal{S} = S(\mathcal{Y})$.
	Let $\psi_{\alpha, \delta, \lambda_0, \lambda_1}^{(D)}, h_{\alpha, \delta, \lambda_0, \lambda_1}^{(D)}$ be the population-level selective prediction-set model for the decision-theoretic approach in Theorem~\ref{thrm:population_version}.
	Let $\psi_{\delta, \lambda_0, \lambda_1}^{(L)}, f_{\delta, \lambda_0, \lambda_1}^{(L)}$ be the population-level selective conditional density model in the robust MLE approach in Theorem~\ref{thrm:population_version}.
	Let population-level prediction-set function associated with the latter model be defined as
	\begin{align}
		h_{\alpha, \delta, \lambda_0, \lambda_1}^{(L)}(x)
		= \argmin_{h \in \mathcal{S(\mathcal{Y})}}
		E_{}\left[
		\ell_\alpha(h, Y)
		\mid X = x
		\right]
		\qquad
		\forall x \in \mathcal{X}'
	\end{align}
	where the expectation is taken with respect to the conditional density $f_{\delta, \lambda_0, \lambda_1}^{(L)}(\cdot |x) \equiv p^*(\cdot|x)$.
	Assume there is a $\beta_0 \ge 0$ and $\beta_1 \in \mathbb{R}$ such that
	\begin{align}
		\ell^*_{\alpha}(X) =
		- \beta_0
		E\left[
		\log p^*(Y|X) \mid X
		\right]
		+ \beta_1
		\qquad
		\forall x \in \mathcal{X}'.
		\label{eq:linear_map}
	\end{align}
	Let $\delta' = \beta_0 \delta + \beta_1$, $\lambda_1' = \lambda_1 \beta_0$, and $\lambda_0' > 0$.
	Then for all $x \in \mathcal{X}'$, we have
	\begin{align}
		\psi_{\delta, \lambda_0, \lambda_1}^{(L)}(x) &= \psi_{\alpha, \delta', \lambda_0', \lambda_1'}^{(D)}(x)\\\
		h_{\alpha, \delta, \lambda_0, \lambda_1}^{(L)}(x) &= h_{\alpha, \delta', \lambda_0', \lambda_1'}^{(D)}(x).
	\end{align}.
	\label{thrm:equiv_models}
\end{theorem}

The following is an example when the linear relation \eqref{eq:linear_map} holds.
\begin{myexample}
	The linear relation \eqref{eq:linear_map} holds when $Y|X=x$ is Gaussian and $\ell_\alpha$ is the absolute discrepancy loss.
	\label{example:gauss}
\end{myexample}
\begin{proof}
	Consider a fixed $x$.
	Let the conditional distribution $Y|X=x$ be $N(\mu, \sigma^2)$.
	The differential entropy of $Y|X=x$ is $\frac{1}{2}\log (2\pi e \sigma^2)$, which is a linear function of $\sigma$.
	The minimum expected absolute discrepancy loss at $x$ is achieved by the interval from the $\alpha/2$ to $1 - \alpha/2$ quantiles of $Y|X=x$.
	Let $Z \sim N(0,1)$.
	The minimum expected absolute discrepancy loss at $x$ is then
	\begin{align}
		\ell^*_{\alpha}(x) & =
		E \left [
		\alpha z_{1 - \alpha/2} \sigma
		+ (\mu - z_{1 - \alpha/2} \sigma - Y)_+
		+ (Y - \mu - z_{1 - \alpha/2} \sigma)_+
		\right]\\
		& =
		\left(
		\alpha z_{1 - \alpha/2}
		+ E \left [
		(z_{1 - \alpha/2} - Z)_+
		+ (Z - z_{1 - \alpha/2})_+
		\right]
		\right)
		\sigma
	\end{align}
	where $z_{\alpha'}$ is the $\alpha/2$ quantile of the standard normal distribution.
	Therefore \eqref{eq:linear_map} is satisfied.
\end{proof}

\section{Implementation details}

\subsection{Example model coupling}
\label{sec:coupling}
It is easy to show that if the decision function is constrained to have the form $\psi(x) = \mathbbm{1}\{ H(x) - \delta > 0 \}$ where $H(x)$ is a lower bound of the expected prediction loss (e.g. entropy in the robust MLE case), the population-level models are the same as that in the unconstrained case.
Thus, we can couple the decision and prediction functions in a similar fashion by defining the decision function as
\begin{align}
	\hat{\psi}(x; \beta)=
	\text{sigmoid}\left(
	\beta (\hat{H}(x) - \delta)
	\right),
	\label{eq:coupled_model}
\end{align}
where $\beta\in \mathbb{R^+}$.
So in the robust MLE approach, we use the estimated entropy for $\hat{H}(x)$; In the decision theoretic approach with  the absolute discrepancy loss, we use $\alpha \hat{r}(x)$ where $\hat{r}(x)$ is the estimated radius.

\subsection{Sampling the uniform acceptance penalty}
\label{sec:pca}
The effect of the uniform acceptance penalty depends on our ability to sample points close to the dataset.
If the training data occupies a negligible volume of $\mathcal{X}$, then the sampled points will generally be far from the training data, and the model will learn to abstain for only far-away regions.
To learn a tighter decision boundary, we apply principal components analysis to the data, and redefine $\mathcal{X}$ to be the values spanned by the transformed training observations.
To avoid aliasing, we can keep all the principal components.
Nevertheless, if a small number of principal components explains most of the variation in the data, then this newly constructed $\mathcal{X}$ will generally have much smaller volume and form a tighter box around the data.
In the empirical analyses, we found that it was sufficient to use only the top principal components that explain > 99\% of the variability.

\section{Coverage guarantees for SPS models}

\begin{proof}[Proof for Theorem~\ref{thrm:bag}]
	First, we prove the result for the conditional coverage rates:
	\begin{align}
		& \E_X\left[
		\left\{
		\Pr\left(
		Y \in H^{\CV}(X) \mid X
		\right)
		- (1 - \alpha)
		\right\}^2
		\mid X \in \tilde{\mathcal{X}}
		\right] \\
		= & \E_X\left[
		\left\{
		\frac{1}{K}
		\sum_{k=1}^K
		\Pr\left(
		Y \in h_k(X) \mid X
		\right)
		- (1 - \alpha)
		\right\}^2
		\mid X \in \tilde{\mathcal{X}}
		\right]\\
		\le & \E_X\left[
		\frac{1}{K}
		\sum_{k=1}^K
		\left\{
		\Pr\left(
		Y \in h_A(X) \mid X, A
		\right)
		- (1 - \alpha)
		\right\}^2
		\mid X \in \tilde{\mathcal{X}}
		\right]\\
		= &
		E_A\left[
		\E_{X}\left[
		\left\{
		\Pr\left(
		Y \in h_{A}(X) \mid X, A
		\right)
		- (1 - \alpha)
		\right\}^2
		\mid X \in \tilde{\mathcal{X}}
		\right]
		\right],
	\end{align}
	where the inequality follows from Jensen's inequality.
	
	Next suppose $\Pr(\psi_k (X) = 1) = \beta$ for all $k = 1,..,K$.
	We now prove the result for marginal coverage rates:
	\begin{align}
		& \left\{ \Pr\left(
		Y \in H^{\CV}(X) \mid \Psi^{\CV}(X) = 1
		\right) - (1 - \alpha)\right\}^2\\
		= &
		\left\{
		\frac{1}{\beta}
		\frac{1}{K} \sum_{k = 1}^K
		\Pr\left(
		Y \in h_k(X) , \psi_k(X) = 1
		\right)
		- (1 - \alpha)\right\}^2\\
		\le &
		\frac{1}{K} \sum_{k = 1}^K
		\left\{
		\frac{1}{\beta}
		\Pr\left(
		Y \in h_k(X) , \psi_k(X) = 1
		\right)
		- (1 - \alpha)\right\}^2\\
		= &
		\E_A\left[
		\left\{
		\Pr\left(
		Y \in h_A(X) \mid \psi_A(X) = 1, A
		\right)
		- (1 - \alpha)\right\}^2
		\right],
	\end{align}
	where the inequality follows from Jensen's inequality.
\end{proof}

Next we prove the asymptotic distribution of the coverage estimator for an ensemble.
We omit the proof for Lemma~\ref{lemma:indiv} since it is very similar.

\begin{proof}[Proof of Theorem~\ref{thrm:asym_normal}]
	Let $\mathbb{P}_0$ denote the expectation with respect to the distribution $P$.
	For a given sample of $n$ observations, let $\mathbb{P}_n$ denote the expectation with respect to the empirical distribution with observations $i = 1,...,n_V$ where the $i$th observation is an array of the $i$th observation in each of the $K$ folds, i.e.
	\begin{align}
		\left(
		\begin{matrix}
			(X_{i,1}, Y_{i,1})\\
			\cdots\\
			(X_{i,K}, Y_{i,K})\\
		\end{matrix}
		\right).
	\end{align}
	Let $\hat{\psi}_{k,n}, \hat{h}_{k,n}$ be the estimated selective prediction-set model for the $k$th fold.
	For $(X, Y)$, define
	$W_{\psi, h} = \psi(X) \mathbbm{1}\left\{
	Y \in h(X)
	\right\}$.
	Then
	\begin{align}
		\left(\mathbb{P}_n - \mathbb{P}_0\right) \left(
		\begin{matrix}
			W_{\hat{\psi}_{1,n}, \hat{h}_{1,n}}\\
			{\hat{\psi}_{1,n}}\\
			\cdots\\
			W_{\hat{\psi}_{K,n}, \hat{h}_{K,n}}\\
			{\hat{\psi}_{K,n}}\\
		\end{matrix}
		\right)
		& =
		\left(\mathbb{P}_n - \mathbb{P}_0\right)
		\left(
		\begin{matrix}
			W_{{\psi}, h}\\
			{\psi}\\
			\cdots\\
			W_{{\psi}, {h}}\\
			{{\psi}}\\
		\end{matrix}
		\right)\\
		& +
		\left(\mathbb{P}_n - \mathbb{P}_0\right)
		\left\{
		\left(
		\begin{matrix}
			W_{\hat{\psi}_{1,n}, \hat{h}_{1,n}}\\
			{\hat{\psi}_{1,n}}\\
			\cdots\\
			W_{\hat{\psi}_{K,n}, \hat{h}_{K,n}}\\
			{\hat{\psi}_{K,n}}\\
		\end{matrix}
		\right)
		-
		\left(
		\begin{matrix}
			W_{{\psi}, h}\\
			{\psi}\\
			\cdots\\
			W_{{\psi}, {h}}\\
			{{\psi}}\\
		\end{matrix}
		\right)
		\right\}.
		\label{eq:remainder}
	\end{align}
	Because the sequence of estimators $(\hat{\psi}_{n}, \hat{h}_{n})$ converges in probability to $({\psi}, {h})$ and the observations are independent and identically distributed, the sequence of estimators for each fold, i.e. $\hat{\psi}_{k,n}, \hat{h}_{k,n}$ for $k = 1,...,K$, converges in probability to $({\psi}, {h})$ in $L_2(P)$.
	It is easy to show that these estimators also jointly converge in probability to $({\psi}, {h})$, i.e. for all $\epsilon > 0$
	\begin{align}
		\Pr\left(
		\left\| \hat{\psi}_{1,n} - \psi \right \|_{P, 2} + \left\| \hat{h}_{1,n} - h \right \|_{P, 2}
		+ ...
		+ \left\| \hat{\psi}_{K,n} - \psi \right \|_{P, 2} + \left\| \hat{h}_{K,n} - h \right \|_{P, 2}
		\ge \epsilon
		\right) \rightarrow 0
	\end{align}
	By the continuous mapping theorem, it follows that
	\begin{align}
		\left(
		\begin{matrix}
			W_{\hat{\psi}_{1,n}, \hat{h}_{1,n}}\\
			{\hat{\psi}_{1,n}}\\
			\cdots\\
			W_{\hat{\psi}_{K,n}, \hat{h}_{K,n}}\\
			{\hat{\psi}_{K,n}}\\
		\end{matrix}
		\right)
		-
		\left(
		\begin{matrix}
			W_{{\psi}, h}\\
			{\psi}\\
			\cdots\\
			W_{{\psi}, {h}}\\
			{{\psi}}\\
		\end{matrix}
		\right)
		\rightarrow_p 0.
	\end{align}
	By the Lindeberg-Feller Central Limit Theorem, \eqref{eq:remainder} is therefore a second-order remainder term where
	\begin{align}
		\sqrt{n_V}
		\left(\mathbb{P}_n - \mathbb{P}_0\right)
		\left\{
		\left(
		\begin{matrix}
			W_{\hat{\psi}_{1,n}, \hat{h}_{1,n}}\\
			{\hat{\psi}_{1,n}}\\
			\cdots\\
			W_{\hat{\psi}_{K,n}, \hat{h}_{K,n}}\\
			{\hat{\psi}_{K,n}}\\
		\end{matrix}
		\right)
		-
		\left(
		\begin{matrix}
			W_{{\psi}, h}\\
			{\psi}\\
			\cdots\\
			W_{{\psi}, {h}}\\
			{{\psi}}\\
		\end{matrix}
		\right)
		\right\}
		\rightarrow_p
		0.
	\end{align}
	
	Let
	\begin{align}
		q &= \E\left[{\psi}(X)\right]\\
		\gamma &= \E\left[{\psi}(X) \mathbbm{1}(Y \in {h}(X)) \right].
	\end{align}
	By the Central Limit Theorem and the Delta Method, it is then easy to show that
	\begin{align}
		\sqrt{n_V}
		\left(
		\frac{\check{\theta}_{\hat{\Psi}_n, \hat{H}_n, n_V} - \theta_{\hat{\Psi}_n, \hat{H}_n}}{\check{\sigma}_{n_V}}
		\right)
		\rightarrow_d
		N \left(
		0,
		\frac{1}{K} {a_0}^\top \Sigma_0 {a_0}
		\right )
	\end{align}
	where $a_0 =	(\begin{matrix} 1/q_0 & -\gamma_0/q_0^2 \end{matrix})$ and $\Sigma_0$ is the covariance matrix of $(W_\psi, \psi)$.
	Finally, by Slutsky's Theorem, we can plug in a consistent estimator of $a_0^\top \Sigma_0 a_0$.
\end{proof}

\subsection{A note on model selection}
\label{sec:model_select}
The main paper only discusses statistical inference for the actual coverage rate of a given SPS model. 
If the coverage rate is too different from the desired rate, one may want to refit the model with a different set of hyperparameters to attain the desired coverage rate.
Here we provide a simple procedure that builds on cSPS to select a model with at least $1 - \alpha$ marginal coverage with high probability and produces confidence intervals for the true coverage of your model.


Our approach is to fit pSPS models for different values of $\alpha$ and run a sequence of hypothesis tests to find the model with the desired coverage rate.
Define a finite monotonically increasing sequence $\{\alpha_j: j = 1,..,J\}$, for which we fit corresponding SPS models $\{(\hat{\psi}_{(j)}, \hat{h}_{(j)}) : j = 1,...,J\}$.
Fix the desired level $\alpha$, and let $\alpha_1 \leq \alpha$ be sufficiently small such that the corresponding model $(\hat{\psi}_{1}, \hat{h}_{1})$ will always have at least $ 1 - \alpha$ marginal coverage.
Next, define a sequence of hypothesis tests $H_0^{(j)}: \theta_{\hat{\psi}_{(j)}, \hat{h}_{(j)}} < 1 - \alpha$ for $j = 1,..,J$.
We test this sequence of hypothesis tests using a gate-keeping procedure, meaning the $j$th hypothesis test is only run if we manage to reject the $(j - 1)$th hypothesis test \citep{Dmitrienko2007-hp}.
Suppose each hypothesis test is performed at fixed level $\tilde{\alpha}$.
The index of the selected model $\hat{g}$ is the largest index for which we successfully reject the null hypothesis.
Because gate-keeping controls the family-wise error rate at $\tilde{\alpha}$, the probability that the selected model has marginal coverage less than $1 - \alpha$ is less than $\tilde{\alpha}$, i.e.
\begin{align}
	\Pr \left (\theta_{\hat{\psi}_{\hat{g}}, \hat{h}_{\hat{g}}} < 1-\alpha \right )
	\le \Pr\left (\text{falsely rejected } H_0^{(j)} \text{ for some } j \in \{1,...,\hat{g}\} \right )
	\le \tilde{\alpha}.
\end{align}

This procedure is based on ideas in \citet{Geifman2017-rl}, which relied on Bonferroni correction and tested all hypotheses at level $\tilde{\alpha}/J$.
However, the Bonferroni correction is known to be overly conservative when the hypothesis tests are highly correlated.
In contrast, the gate-keeping procedure is a simple solution that leverages the natural ordering of the tests, in that models fit with smaller $\alpha_j$ likely have higher marginal coverage rates.

\section{Experimental details}
The neural networks in all of our experiments are densely connected unless specified otherwise.
The models were all implemented in Tensorflow \citep{Abadi2016-zh}.

\subsection{Simulation settings}
\label{sec:sim_settings}

\begin{table}
	\begin{tabular}{c|c}
		Simulation & Conditional Distribution $Y|X= x $ \\
		\toprule
		Fig~\ref{fig:sims}A & $N \left((|0.5 x_1 + x_2| + |x_1| - 0.5 | x_2 |, (0.08 (|x_1| + 3) + |x_2 - 3|  + 0.1)^2 \right)$\\
		\midrule
		Fig~\ref{fig:sims}B & $N \left((|0.5 x_1 + x_2| + |x_1| - 0.5 | x_2 |, 1 \right)$\\
		\midrule
		Fig~\ref{fig:sims}C &
		$
		\begin{cases}
			N\left(|0.5 x_1 + x_2| + |x_1| - 0.5 | x_2 |, 0.09\right)
			& \text{if } \|x\|_{\infty} > 1
			\\
			N\left( x_1^2 x_2^2, 0.09\right) & \text{if } \|x\|_{\infty} \le 1
		\end{cases}
		$\\
		\midrule
		Fig~\ref{fig:sim_coverage} & $N\left(
		\tanh(x_1 + 2x_2) + x_1 + \tanh(x_3 + 2x_4),
		(1.1 + \tanh(x_1 + 2x_2))^2
		\right )$\\
		\midrule
		Fig~\ref{fig:compare} & N$\left(
		\sin(x_1 \mathbbm{1}\{x_1 > 0\}
		+ 0.5x_1
		+ 0.9 (x_1 + 4) \mathbbm{1}\{x_1 < -4\}
		) - 0.25 (x_1 - 5) \mathbbm{1}\{x_1 \ge 5\},0.01
		\right)$
	\end{tabular}
	\caption{Simulation settings: conditional distributions}
	\label{table:sim_settings}
\end{table}

See Table~\ref{table:sim_settings} for simulation settings.
For the simulations evaluating confidence interval coverage and widths, we used 200 replicates.
To fit the models, we considered neural networks with one to three hidden layers with ten to twenty hidden nodes per layer.
The network structures were pretuned using cross-validation.

\subsection{Additional Simulations}
\label{sec:addition_results}

In this section, we present additional simulation results.

\begin{figure}
		\begin{tabular}{cp{0.8\linewidth}}
			\raisebox{-.8\height}{\includegraphics[width=0.2\linewidth]{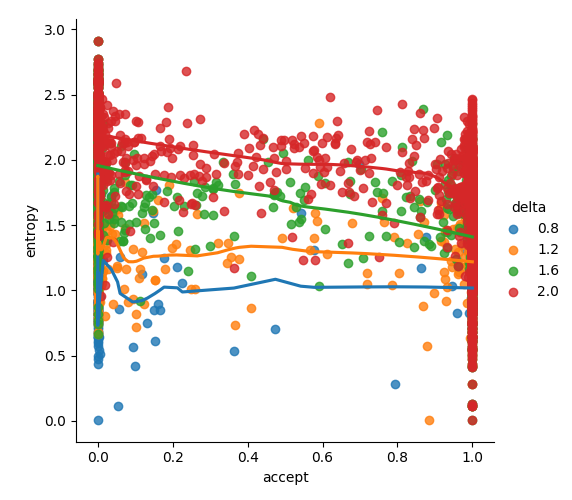}}
			& A) Plot of the acceptance probabilities for pSPS models fit using the abstention costs $\delta = 0.8, 1.2, 1.6, 2.0$ versus the true entropy of $Y|X = x$.
			The data was $p=10$ dimensional and the model was trained on 20000 observations.
			\\
			\midrule
			\raisebox{-.8\height}{\includegraphics[width=0.2\linewidth]{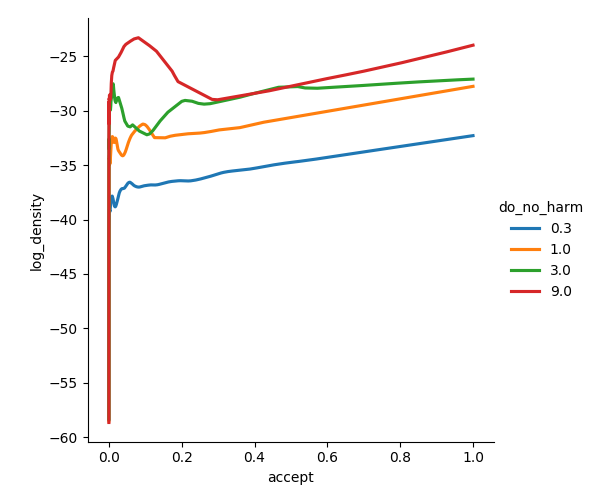}} &
			B) Plot of the acceptance probabilities for pSPS models fit using penalty parameters $\lambda_1 = 0.3, 1, 3.0, 9.0$ versus the true density of $x$.
			The data was $p=10$ dimensional and the model was trained on 20000 observations.
			\\
			\midrule
			\raisebox{-.8\height}{\includegraphics[width=0.2\linewidth]{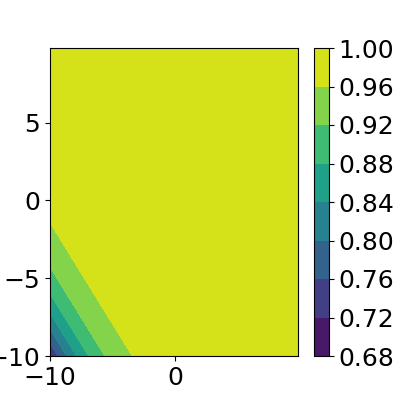}} &
			C) The acceptance probability for a pSPS model over the domain $\mathcal{X} = [-10,10]^2$.
			Here, the true conditional mean of $Y|X=x$ is linear everywhere but over the region $[-1,1]^2$.
			Decision and prediction model are restricted to be linear models.
		\end{tabular}
	\caption{Simulation results for penalized loss minimization for SPS models. This figure appears in color in the electronic version of this article.}
	\label{fig:sims_additional}
\end{figure}

First, we ran simulations analogous to those in Section~\ref{sec:simulations} in the main manuscript but for higher dimensional data.
Figure~\ref{fig:sims_additional} A and B are analogous to Figure~\ref{fig:sims} A and B of the main manuscript, except with $p=10$ variables and 20000 observations.
Figure~\ref{fig:sims_additional}A plots the true conditional entropy of $Y|X$ versus the acceptance probability estimated by pSPS.
By varying the value of the abstention cost $\delta$, we see that the decision function approximately thresholds on the true conditional entropy.
Figure~\ref{fig:sims_additional}B plots the true density $p^*(x)$ at test points $x$ versus the acceptance probability estimated by pSPS.
By varying the value of the penalty parameter $\lambda_1$ for the uniform acceptance penalty, the decision function approximately thresholds on different densities.
We note that the estimated decision functions are not perfect discriminators between points above and below the entropy or density cutoff, and improve as the training dataset size increases.
Thus, these empirical results reflect the findings in Theorem~\ref{thrm:population_version}, but show that higher-dimensional data settings require larger number of observations recover the asymptotic properties.

Recall that Figure~\ref{fig:sims}C of the main manuscript showed that a nonparametric decision function trained in conjunction with a highly parametric prediction model protects against model misspecification.
Parametric decision functions do not necessarily offer this same protection and may, in fact, convey a false sense of safety.
To illustrate this fact, we ran the same experiment as before, where $\E[Y|X=x]$ is linear in $x$ over $\mathcal{X} = [-10,10]^2$, except in $[-1,1]^2$ where it follows a quadratic function.
We fit a pSPS model where both the prediction and decision functions are linear in $x$.
As shown in Figure~\ref{fig:sims_additional}C, the fitted decision function accepts with 100\% over most of $\mathcal{X}$ and is less confident for points close to (-10,-10).
The decision function is therefore under-confident in regions where the linear model is correct and over-confident in regions where the linear model is inappropriate.
As such, we suggest selecting flexible model classes for the decision function.

\subsection{MNIST: Additional details}
\label{sec:mnist_appendix}
We performed PCA on the images of digits zero to eight and kept the top 300 principal components, which explained over 99\% of the variation.
The first approach trained dense neural networks with two hidden layers to predict digits given PC scores.
The second approach trained convolutional neural networks with the following hidden layers: a $32\times 3\times 3$ convolutional layer, a $2\times 2$ max-pooling layer, a $32 \times 3\times 3$ convolutional layer, a $2\times 2$ max-pooling layer.
For the uniform acceptance penalty, $\mathcal{X}$ was defined as the smallest box containing the PC scores on the training data in the first approach; $\mathcal{X}$ was the inverse PC transform of the smallest box containing the PC scores in the second approach.
The displayed results are from 20 evaluation runs, where each run trained models using 54000 randomly selected images.

\subsection{ICU prediction tasks: Additional details}
\label{sec:mimic_appendix}
\begin{figure}[!p]
	\centering
	\includegraphics[height=0.14\linewidth]{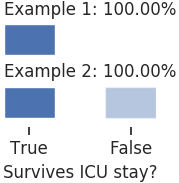}
	\includegraphics[height=0.14\linewidth]{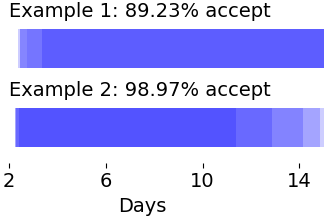}
	\caption{
		In-hospital mortality prediction sets (left) and length-of-stay prediction intervals (right) for example subjects, with acceptance probabilities.
		Prediction interval/sets from the fitted models from each fold are overlaid on top of each other; dark regions indicate more consensus.
		This figure appears in color in the electronic version of this article.
	}
	\label{fig:example_mimic_predictions}
\end{figure}

For the ICU data analyses on the MIMIC dataset, we followed the data cleaning and feature extraction procedure in \cite{Harutyunyan2017-hw}, which outputs age, gender, ethnicity, and 42 summary statistics for the time series of 17 physiological measurements over the first 48 hours of admission.
The full list of clinical variables, which includes measurements like blood pressure, temperature, and glucose levels, are given in Table~1 in \cite{Harutyunyan2017-hw}.
We remove all features where more than 10\% of the data was missing and the final dataset contained 297 variables.
(We note that our analysis includes all patients, whereas their analysis removes neonates.)
We then ran PCA on the continuous covariates and we kept the top 120 principal components, as these explained over 99\% of the variation.
The displayed results are from 20 evaluation runs, where the training data was composed of roughly 3900 randomly selected observations.
Example predictions from the ensembles are shown in Figure~\ref{fig:example_mimic_predictions}.

\subsection{Computation time}
For the experiments in this paper, all the models could be fit in a couple of minutes.
The computation time for fitting penalized selective prediction set models was only slightly longer than fitting ordinary prediction set models.
There are two main reasons for this increased computation time.
First, if one wants to fit a separate neural network for the decision function, then there is additional hyperparameter tuning.
However, if the prediction model is already a neural network, this can be entirely avoided using the coupling procedure (see Section~\ref{sec:coupling}).
Second, one needs to uniformly sample points from $\mathcal{X}$ to approximate the uniform acceptance penalty.
However, this is both easy and quick to do, so the increase in training time from this is negligible.

\section{Choosing the number of folds $K$}
\label{sec:kfolds_choice}
In Section~\ref{sec:bag_recalib}, we highlighted the benefits of ensembling using a $K$-fold cross-validation procedure.
In particular, we showed that the confidence intervals for the marginal coverage were narrower than the individual models and the coverage was closer to the nominal rate.
Here, we study how the number of folds impacts the fitted model and statistical inference of its coverage rates.
We use the same settings as that used in Fig~\ref{fig:sim_coverage}, but fix the number of training observations to 720 and vary $K = 3,5, 10, 20$.

We find that $K$ has relatively little effect on the coverage and width of the confidence interval for the ensemble.
This is because there are two factors at play when we vary $K$ and they actually cancel out.
First, the asymptotic variance for the coverage estimators of the $K$ individual models increases at the rate of $K$.
Second, the asymptotic variance of the ensemble is $1/K$ of that for individual models.
When multiplied, the asymptotic variance of the ensemble does not depend on $K$, at least asymptotically in $n$.
There does seem to be a slight advantage in that the conditional coverage rates of the ensemble for higher $K$ tends to be closer to the nominal rate of 80\%.
This is primarily because the training data for the individual models become more similar as $K$ increases.
However, if $K$ is too large, the number of validation samples becomes too small for asymptotics to apply.

From this simulation study, we conclude that the choice of $K$ has a relatively small effect on the trained model and its coverage estimates and confidence intervals.
In practice, this means that one could simply use popular values for $K$ when evaluating model performance, such as $3, 5$, and $10$.
There also seems to be marginal gains for using slightly larger $K$, as long as the validation sets are sufficiently large for asymptotic normality to hold.

\begin{figure}
	\includegraphics[width=\linewidth]{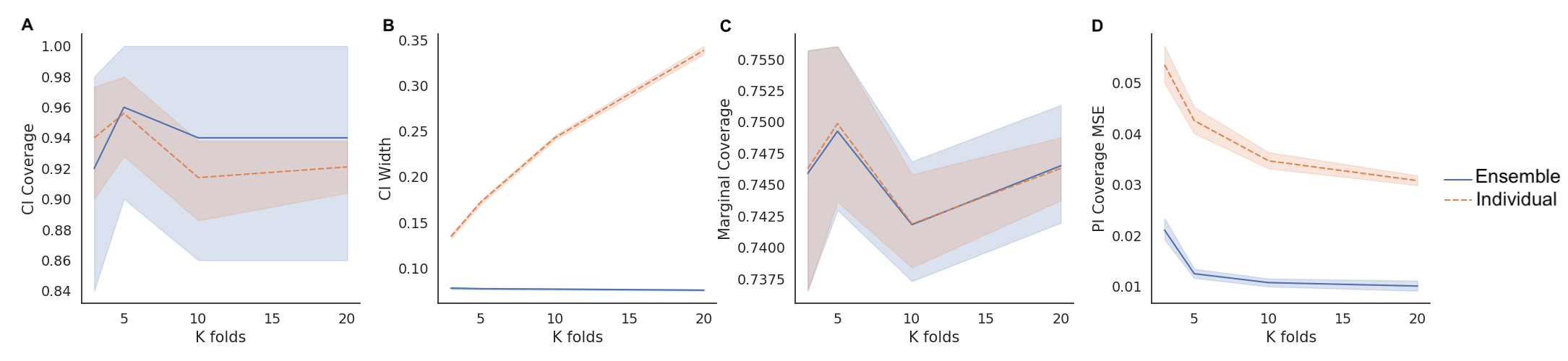}
	\caption{
		Coverage results when varying the number of folds $K = 3,5,10,20$.
		(A) The CIs achieve the desired 95\% coverage rates.
		(B) The CI width for the ensemble's coverage is essentially constant in $K$.
		(C) The marginal coverage of the models are also essentially constant in $K$.
		(D) The mean squared error (MSE) of the ensemble's conditional coverage from the nominal rate (for accepted $x$) decreases in $K$ but quickly levels off.
		This figure appears in color in the electronic version of this article.
	}
	\label{fig:k_folds}
\end{figure}

\end{document}